%% file: arxiv.tex
\documentclass[nohyperref]{article}

\usepackage{microtype}
\usepackage{graphicx}
\usepackage{subfigure}
\usepackage{booktabs} % for professional tables
\usepackage{overpic}
\usepackage{caption}
\usepackage{hyperref}

\usepackage[accepted]{icml2022}

\usepackage{amsmath}
\usepackage{amssymb}
\usepackage{mathtools}
\usepackage{amsthm}
\usepackage{multirow}
\usepackage{footnote}
\usepackage{enumitem}
\usepackage{graphicx,animate} % for \animategraphics
\usepackage{bm}
\usepackage{grffile}

\usepackage{multimedia}
\usepackage[english]{babel}
\usepackage{media9} % for \includemedia AVOID as about to expire
\usepackage[capitalize,noabbrev]{cleveref}

\theoremstyle{plain}
\newtheorem{theorem}{Theorem}[section]
\newtheorem{proposition}[theorem]{Proposition}

\theoremstyle{definition}

\theoremstyle{remark}

\def \pzo {\phantom{0}}

\usepackage[textsize=tiny]{todonotes}

\icmltitlerunning{Understanding The Robustness in Vision Transformers}

\begin{document}

\twocolumn[
\icmltitle{Understanding The Robustness in Vision Transformers}
\icmlsetsymbol{intern}{*}
\icmlsetsymbol{affil}{\dag}

\begin{icmlauthorlist}
\icmlauthor{Daquan Zhou}{nus,intern}
\icmlauthor{Zhiding Yu}{nv}
\icmlauthor{Enze Xie}{hku}\\
\icmlauthor{Chaowei Xiao}{nv,asu}
\icmlauthor{Anima Anandkumar}{nv,caltech}
\icmlauthor{Jiashi Feng}{bytedance,affil}
\icmlauthor{Jose M. Alvarez}{nv}
\end{icmlauthorlist}

\icmlaffiliation{nus}{National University of Singapore}
\icmlaffiliation{nv}{NVIDIA}
\icmlaffiliation{hku}{The University of Hong Kong}
\icmlaffiliation{asu}{ASU}
\icmlaffiliation{caltech}{Caltech}
\icmlaffiliation{bytedance}{ByteDance}

\icmlcorrespondingauthor{Zhiding Yu}{zhidingy@nvidia.com}

\vskip 0.3in
]

\printAffiliationsAndNotice{\icmlIntern} % otherwise use the standard text.

\begin{abstract}
Recent studies show that Vision Transformers (ViTs) exhibit strong robustness against various corruptions. Although this property is partly attributed to the self-attention mechanism, there is still a lack of systematic understanding. In this paper, we examine the role of self-attention in learning robust representations. Our study is motivated by the intriguing properties of the emerging visual grouping  in Vision Transformers, which indicates that self-attention may promote robustness through improved mid-level representations. We further propose a family of fully attentional networks (FANs) that strengthen this capability by incorporating an attentional channel processing design. We validate the design comprehensively on various hierarchical backbones. Our model achieves a state-of-the-art 87.1\% accuracy and 35.8\% mCE on ImageNet-1k and ImageNet-C with 76.8M parameters. We also demonstrate state-of-the-art accuracy and robustness in two downstream tasks: semantic segmentation and object detection. Code is available at \url{https://github.com/NVlabs/FAN}.
\end{abstract}

\input{tex/intro}
\input{tex/method}
\input{tex/experiment}
\input{tex/related}

\section{Conclusion}
In this paper, we verified self-attention as a contributor of the improved robustness in vision transformers. 
Our study shows that self-attention promotes naturally formed clusters in tokens, which exhibits interesting relation to the extensive early studies in vision grouping prior to deep learning. We also established an explanatory framework from the perspective of information bottleneck to explain these  properties of self-attention. To push the boundary of robust representation learning with self-attention, we introduced a family of fully-attentional network (FAN) architectures, where self-attention is leveraged in both token mixing and channel processing. FAN models demonstrate significantly improved robustness over their CNN and ViT counterparts. Our work provides a new angle towards understanding the working mechanism of vision transformers, showing the potential of inductive biases going beyond convolutions. Our work can benefit wide real-world applications, especially safety-critical ones such as autonomous driving.
\bibliography{ref}
\bibliographystyle{icml2022}

\input{tex/appendix}

\end{document}

%% file: tex/intro.tex
\section{Introduction}
\label{sec:intro}
\begin{figure}[!ht] 
\small
\centering
    \begin{minipage}[c]{\linewidth}
    \tiny
    \begin{overpic}[width=\textwidth]{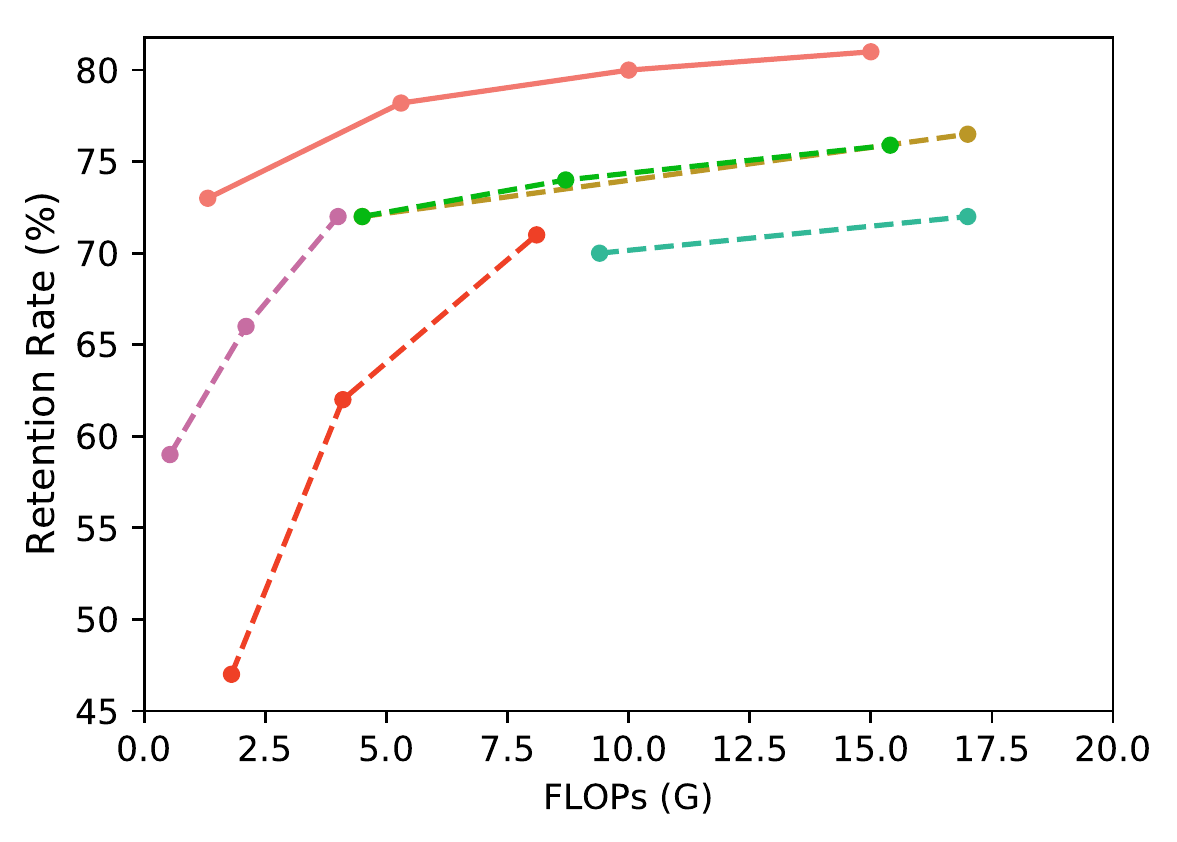}
    \put(75.5,65.9){FAN (Ours)}
    \put(83.3,59.5){DeiT}
    \put(77,57.2){ConvNeXt}
    \put(82.5,52.5){ViT}
    \put(30.5,51.1){PVT\_V2}
    \put(46.5,51.5){ResNet}
    \put(41,26){
    \tiny
    \setlength\tabcolsep{0.5mm}
    \renewcommand{\arraystretch}{1}
    \begin{tabular}{l|ccc} 
      Model  & \#Param. & Clean / Robust  \\ \hline
      Res18 (\citeauthor{he2016deep})  & 11M  & 69.0 / 32.7 \\
      FAN-T-ViT (Ours) & ~7M & 79.2 / 54.2 \\ \hline
      Res50 (\citeauthor{he2016deep})  & 25M  & 79.0 / 50.6\\
      DeiT-S (\citeauthor{touvron2021training})  & 22M  & 79.9 / 58.1 \\
      FAN-S-ViT (Ours) & 28M & 82.6 / 64.5 \\ \hline
      Res101 (\citeauthor{he2019bag}) &45M  & 83.0 / 59.2 \\
      DeiT-B (\citeauthor{touvron2021training}) &89M &  82.0 / 62.8 \\ 
      FAN-B-ViT (Ours) & 54M  & 83.6 / 67.0\\ 
      FAN-L-Hybrid (Ours) & 77M  & 84.3 / 68.3\\ 
    \end{tabular}}
    \end{overpic}
    \end{minipage}\hfill
    \begin{minipage}{0.328\linewidth}
        \centering
        \includegraphics[width=\textwidth]{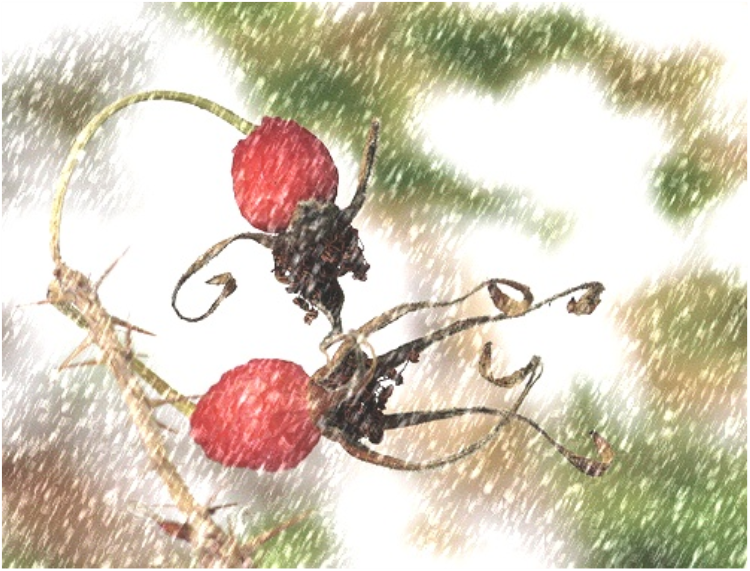}\\
        Corrupted input
    \end{minipage}\hfill
    \begin{minipage}{0.328\linewidth}
        \centering
        \includegraphics[width=\textwidth]{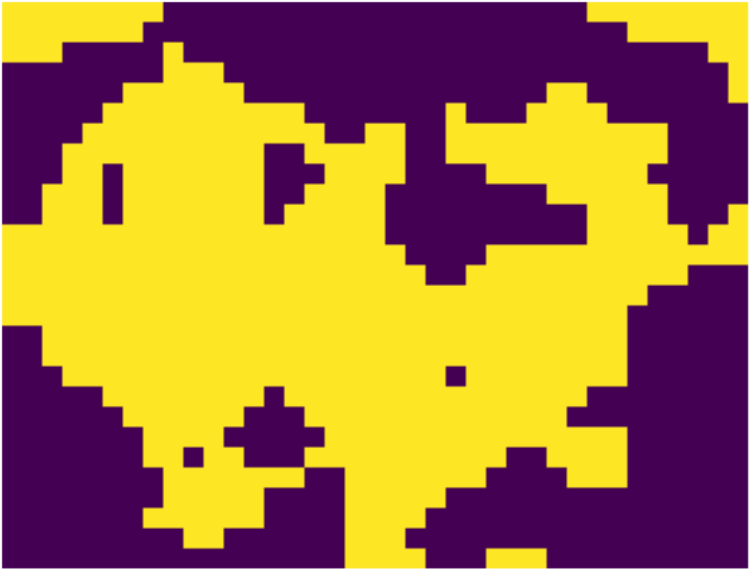}\\
        ResNet-50
    \end{minipage}\hfill
    \begin{minipage}{0.328\linewidth}
        \centering
        \includegraphics[width=\textwidth]{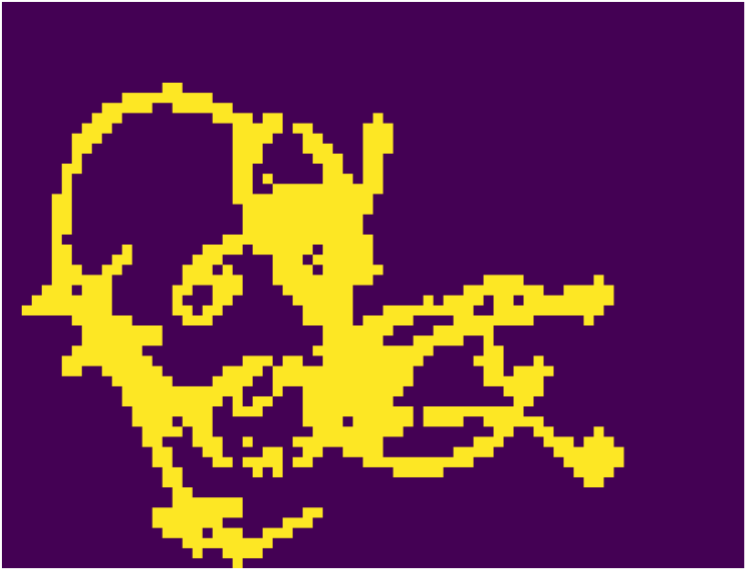}\\
        FAN-S (ours)
    \end{minipage}
\caption{\textbf{Main results on ImageNet-C (top figure) and clustering visualization (bottom row).} Retention rate is defined as robust accuracy / clean accuracy. Left to right in bottom row: input image contaminated by corruption (snow) and the visualized clusters. Visualization is conducted on the output features (tokens) of the second last layers. All models are pretrained on ImageNet-1K. Input size is set to $448\times 448$ following~\cite{caron2021emerging}. 
}
\label{fig:teaser}
\end{figure}

Recent advances in visual recognition are marked by the rise of Vision Transformers (ViTs)~\cite{dosovitskiy2020image} as state-of-the-art models. Unlike ConvNets~\cite{lecun1989backpropagation,krizhevsky2012imagenet} that use a ``sliding window'' strategy to process visual inputs, the initial ViTs feature a design that mimics the Transformers in natural language processing - An input image is first divided into a sequence of patches (tokens), followed by self-attention (SA)~\cite{vaswani2017attention} layers to aggregate the tokens and produce their representations. Since introduction, ViTs have achieved good performance in many visual recognition tasks.

Unlike ConvNets, ViTs incorporate the modeling of non-local relations using self-attention, giving it an advantage in several ways. 
An important one is the robustness against various corruptions. Unlike standard recognition tasks on clean images, several works show that ViTs consistently outperform ConvNets by significant margins on corruption robustness~\cite{bai2021transformers,xie2021segformer,paul2022vision,naseer2021intriguing}. The strong robustness in ViTs is partly attributed to their self-attention designs, but this hypothesis is recently challenged by an emerging work ConvNeXt~\cite{liu2022convnet}, where a network constructed from standard ConvNet modules without self-attention competes favorably against ViTs in generalization and robustness. This raises an interesting question on the actual role of self-attention in robust generalization.

\textbf{Our approach:} In this paper, we aim to find an answer to the above question. Our journey begins with the intriguing observation that meaningful segmentation of objects naturally emerge in ViTs during image classification~\cite{caron2021emerging}. This motivates us to wonder whether self-attention promotes improved mid-level representations (and thus robustness) via visual grouping - a hypothesis that echoes the odyssey of early computer vision~\cite{grouping}. As a further examination, we analyze the output tokens from each ViT layer using spectral clustering~\cite{ng2002spectral}, where the significant\footnote{eigenvalues are larger than a predefined threshold $\epsilon$.} eigenvalues of the affinity matrix correspond to the main cluster components. Our study shows an interesting correlation between the number of significant eigenvalues and the perturbation from input corruptions: both of them decrease significantly over mid-level layers, which indicates the symbiosis of grouping and robustness over these layers.

\begin{figure}[t] 
\centering
\includegraphics[width=1\linewidth]{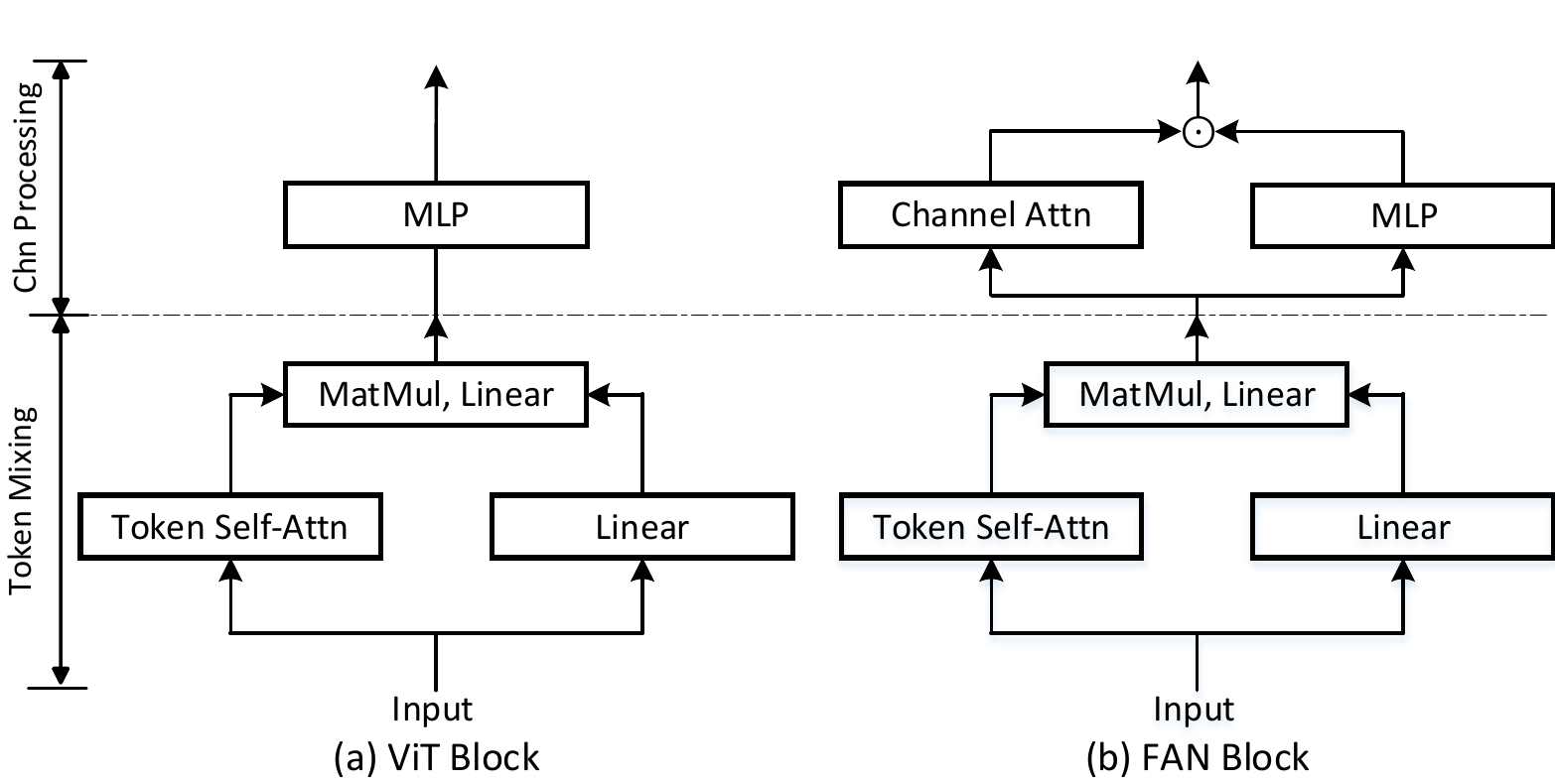}
\caption{\textbf{Comparison between conventional ViT block and the proposed FAN block.} (a) ViT block: Input tokens are first aggregated by self-attention, followed by a linear projection and an MLP is appended to the self attention block for feature transformation. (b) FAN block: both token self-attention and channel attention are applied, which makes the entire network fully attentional. The linear projection layer after the channel attention is removed.
}
\vspace{-7mm}
\label{fig:fan_arch}
\end{figure}

To understand the underlying reason for the grouping phenomenon, we interpret SA from the perspective of information bottleneck (IB)~\cite{tishby2000information,tishby2015deep}, a compression process that ``squeezes out'' unimportant information by minimizing the mutual information between the latent feature representation and the target class labels, while maximizing mutual information between the latent features and the input raw data. We show that under mild assumptions, self-attention can be written as an iterative optimization step of the IB objective. This partly explains the emerging grouping phenomenon since IB is known to promote clustered codes~\cite{still2003geometric}.

As shown in Fig.\ref{fig:fan_arch} (a), previous Vision Transformers often adopt a multi-head attention design, followed by an MLP block to aggregate the information from multiple separate heads. Since different heads tend to focus on different components of objects, the multi-head attention design essentially forms a mixture of information bottlenecks. As a result, how to aggregate the information from different heads matters. We aim to come up with an aggregation design that strengthens the symbiosis of grouping and robustness. As shown in Fig.\ref{fig:fan_arch} (b), we propose a novel attentional channel processing design which promotes channel selection through reweighting. Unlike the static convolution operations in the MLP block, the attentional design is dynamic and content-dependent, leading to more compositional and robust representations. The proposed module results in a new family of Transformer backbone, coined Fully Attentional Networks (FANs) after their designs.

Our contributions can be summarized as follows:
\begin{itemize}[leftmargin=1.3em,topsep=1pt,itemsep=0.1pt]
\item Instead of focusing on empirical studies, this work provides an explanatory framework that unifies the trinity of grouping, information bottleneck and robust generalization in Vision Transfomrers.
\item The proposed fully attentional design is both efficient and effective, bringing systematically improved robustness with marginal extra costs. Compared with state-of-the-art architectures such as ConvNeXt, our model shows favorable performance in both clean and robust accuracy in image classification. For instance, our model achieves 47.7\% mCE on ImageNet-C with 28M parameters, better than ResNet-50, Swin-T and recent SOTA ConvNeXt-T by 29.0\%, 11.9\% and 5.5\% under the comparable model size. By scaling the FAN model to 76.8M model size, we achieve 35.8\% mCE, the new state-of-the-art robustness under all supervised trained models. 
\item We also conduct extensive experiments in semantic segmentation and object detection. We show that the significant gain in robustness from our proposed design is transferrable to these downstream tasks.
\end{itemize}

Our study indicates the non-trivial benefit of attention representations in robust generalization, and is in line with the recent line of research observing the intriguing robustness in ViTs. We hope our observations and discussions can lead to a better understanding of the representation learning in ViTs and encourage the community to go beyond standard recognition tasks on clean images.

%% file: tex/method.tex
\section{Fully Attentional Networks}

In this section, we examine some emerging properties in ViTs and interpret these properties from an information bottleneck perspective. We then present the proposed Fully Attentional Networks (FANs).

\subsection{Preliminaries on Vision Transformers}
A standard ViT  first divides an input image into $n$  patches uniformly and
encodes each patch into a token embedding $\mathbf{x}_i \in \mathbb{R}^d, i=1,\ldots,n$. 
Then, all these tokens  are fed into a stack of transformer blocks. 
Each transformer block leverages self-attention for token mixing and MLPs for channel-wise feature transformation. The architecture of a transformer block is illustrated in the left of Figure \ref{fig:fan_arch}.

\textbf{Token mixing.} Vision transformers leverage self-attention to aggregate global information. Suppose the input token embedding tensor is $X =[\mathbf{x}_1, \ldots, \mathbf{x}_n]\in\mathbb{R}^{ d \times n }$, SA applies linear transformation with parameters $W_K, W_Q, W_V$ to embed them into the key $K=W_K X \in \mathbb{R}^{d\times n}$, query $Q=W_Q X \in \mathbb{R}^{d\times n}$ and value $V = W_V X \in \mathbb{R}^{d\times n}$ respectively. The SA module then computes the   attention  matrix and aggregates the token features as follows:

\vspace{-0.2cm}
\begin{small}
\begin{equation}
\label{eqn:vanilla_sa}
     Z^\top  = \mathrm{SA}(X) = \mathrm{Softmax}\left(\frac{Q^\top K}{\sqrt{d}}\right) V^\top W_L, 
\end{equation}
\end{small}

\vspace{-0.3cm}
where $W_L\in \mathbb{R}^{d\times d}$ is a linear transformation and $Z = [\mathbf{z}_1, \ldots, \mathbf{z}_n]$ is the aggregated token features and $\sqrt{d}$ is a scaling factor. The output of the SA is then normalized and   fed into the MLP to generate the input to the next block. 

\textbf{Channel processing.} Most ViTs adopt an MLP block to transform the input tokens into features $Z$:

\vspace{-0.2cm}
\begin{small}
\begin{equation}
    \label{eqn:mlp}
    Z' = \mathrm{MLP}(Z).
\end{equation}
\end{small}

\vspace{-0.5cm}
The block contains two Linear layers and a GELU layer.

\subsection{Intriguing Properties of Self-Attention}\label{subsec:sa_property}

\begin{figure*}[t] 
	\small
	\centering
	\begin{minipage}{0.328\linewidth}
		\centering
		\includegraphics[width=\textwidth]{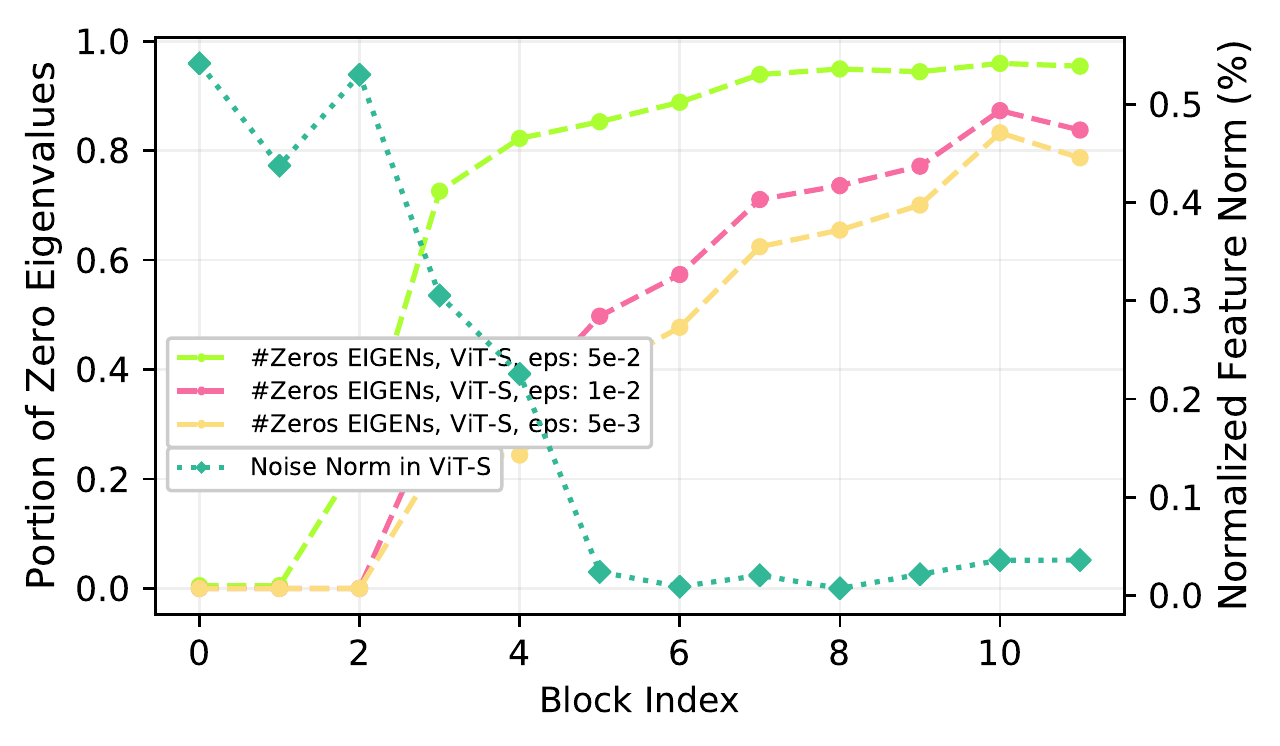}\\ \vspace{-2mm}
		(a)
	\end{minipage}\hfill
	\begin{minipage}{0.328\linewidth}
		\centering
		\includegraphics[width=\textwidth]{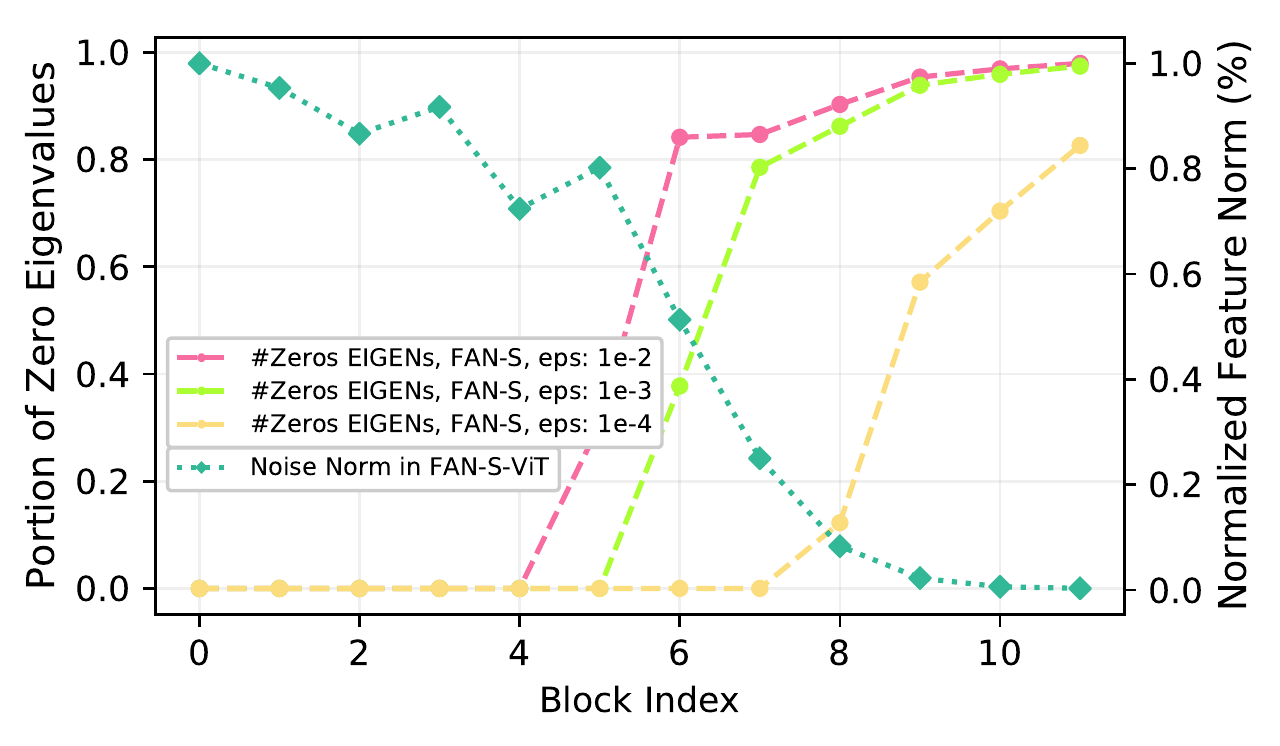}\\ \vspace{-2mm}
		(b)
	\end{minipage}\hfill
	\begin{minipage}{0.302\linewidth}
		\centering
		\includegraphics[width=\textwidth]{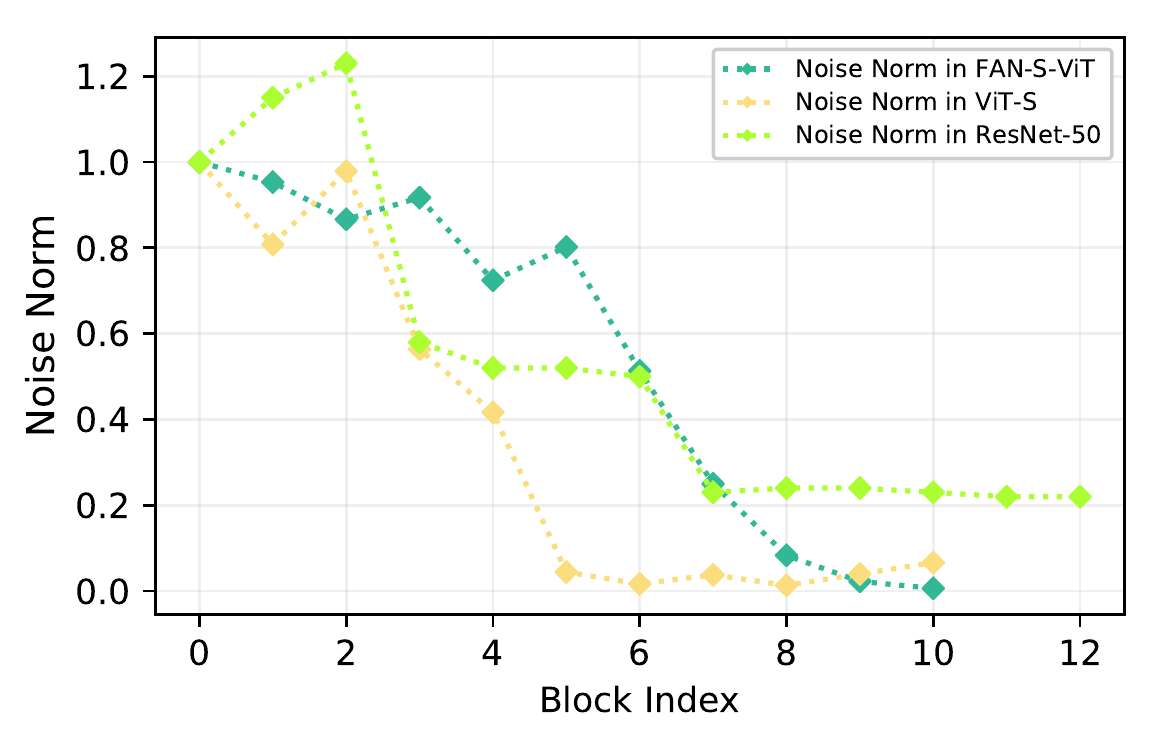}\\ \vspace{-2mm}
		(c)
	\end{minipage}
	\vspace{-3mm}
	\caption{\textbf{Analysis on the grouping of tokens and noise decay.} (a) and (b) shows the \# of insignificant (zero) eigenvalues and the noise input decay of ViT-S and FAN-S respectively; (c) shows the comparison of noise norm across different blocks in FAN-S, ViT-S and ResNet-50. Plots shown in (a) and (b) show that the number of zero eigenvalues increases as the model goes deeper, which indicates the emerging grouping of tokens. Given the input Gaussian noise, its magnitude similarly decays over more self-attention blocks. Such a phenomenon is not observed in the ResNet-50 model.
	}
	\label{fig:clustering_number}
\end{figure*}

We begin with the observation that meaningful clusters emerge on ViT's token features $\mathbf{z}$. We examine such phenomenon using spectral clustering~\cite{ng2002spectral}, where the token affinity matrix is defined as $S_{ij} = \mathbf{z}_i^\top \mathbf{z}_j$. Since the number of major clusters can be estimated by the multiplicity of significant eigenvalues~\cite{manor2004self} of $S$, we plot the number of (in)significant eigenvalues across different ViT-S blocks (Figure~\ref{fig:clustering_number} (a)). We observe that by feeding Gaussian noise $\mathbf{x} \sim \mathcal{N}(0,1)$, the resulting perturbation (measured the by normalized feature norm) decreases rapidly together with the number of significant eigenvalues. Such observation indicates the symbiosis of grouping and improved robustness over middle blocks. 

\begin{figure}[t] 
	\centering
	\includegraphics[width=1\linewidth]{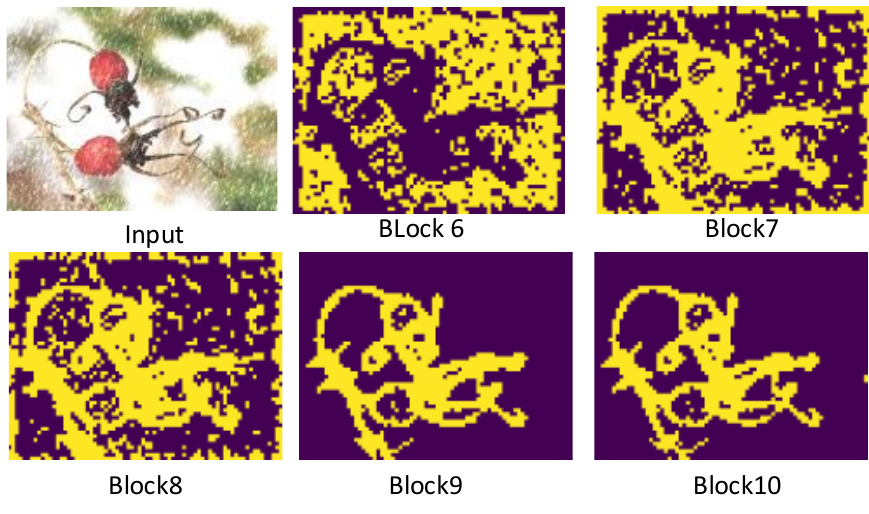}
	\vspace{-6mm}
	\caption{\textbf{Clustering visualization for different blocks.} The visualization is based on our proposed FAN-S model as detailed in  Table \ref{tab:model_arch}. The cluster visualizations are generated by applying spectral clustering on token features from each FAN block.}
	\label{fig:clustering_visualization}
	\vspace{-3mm}
\end{figure}

We additionally visualize the same plot for FAN-S-ViT in Figure~\ref{fig:clustering_number} (b) where similar trend holds even more obviously. The noise decay of ViT and FAN is further compared to ResNet-50 in Figure~\ref{fig:clustering_number} (c). We observe that: 1) the robustness of ResNet-50 tends to improve upon downsampling but plateaus over regular convolution blocks. 2) The final noise decay of ResNet-50 less significant. Finally, we visualize the grouped tokens obtained at different blocks in Figure~\ref{fig:clustering_visualization}, which demonstrates the process of visual grouping by gradually squeezing out unimportant components. 
Additional visualizations on different features (tokens) from different backbones are provided in the appendix.

\subsection{An Information Bottleneck Perspective}
\label{sec:explain}

The emergence of clusters and its symbiosis with robustness in Vision Transformers draw our attention to early pioneer works in visual grouping~\cite{grouping,buhmann1999image}. In some sense, visual grouping can also be regarded as some form of lossy compression~\cite{yang2008unsupervised}. We thus present the following explanatory framework from an information bottleneck perspective. 

Given a  distribution $X \sim \mathcal{N}(X',\epsilon)$ with $X$ being the observed noisy input and $X'$ the target clean code, IB seeks a mapping $f(Z|X)$ such that $Z$ contains the relevant information in $X$ for predicting $X'$. This goal is formulated as the following information-theoretic optimization problem:

\vspace{-0.1cm}
\begin{small}
\begin{equation}
\label{eqn: ib_objective}
    f^*_{\mathrm{IB}}(Z|X) = \arg\min_{f(Z|X)} I(X,Z) - I(Z,X'), 
\end{equation}
\end{small}

\vspace{-0.3cm}
Here the first term compresses the information  and the second term encourages to maintain the  relevant information. 

In the case of an SA block, $Z = [\mathbf{z}_1, \ldots, \mathbf{z}_n]  \in \mathbb{R}^{ d \times n }$ denote the output features and $X = [\mathbf{x}_1, \ldots, \mathbf{x}_n]  \in \mathbb{R}^{ d \times n }$ the input. 
Assuming $i$ is the data point index, we have:

\begin{proposition}
\label{prop:ib_sa}
Under mild assumptions, the iterative step to optimize the objective in Eqn.~\eqref{eqn: ib_objective} can be written as:

\vspace{-0.2cm}
\begin{small}
\begin{equation}
    \mathbf{z}_c = \sum_{i=1}^n \frac{\log [n_c / n] }{n \det \Sigma} \frac{\exp\left[ \frac{\mu_c^\top \Sigma^{-1} \mathbf{x}_i}{1/2}   \right]}{ \sum_{c=1}^n \exp\left[ \frac{\mu_c^\top \Sigma^{-1} \mathbf{x}_i}{1/2}  \right]} \mathbf{x}_i,
\end{equation}
\end{small}

\vspace{-0.3cm}
or in matrix form:

\vspace{-0.2cm}
\begin{small}
\begin{equation}
    Z = \mathrm{Softmax}(Q^\top K /d ) V^\top,
\end{equation}
\end{small}

\vspace{-0.3cm}
with $V =[\mathbf{x}_1,\ldots,\mathbf{x}_N] \frac{\log [n_c/n]}{n \det \Sigma}$, $K =  [\mu_1, \ldots, \mu_N] = W_K X$, $Q = \Sigma^{-1} [\mathbf{x}_1,\ldots, \mathbf{x}_N]$ and $d=1/2$. Here $n_c$, $\Sigma$ and $W_K$ are  learnable variables. 
\end{proposition}

\textbf{Remark.} We defer the proof to the appendix. The above proposition establishes an interesting connection between the vanilla self-attention \eqref{eqn:vanilla_sa} and IB \eqref{eqn: ib_objective}, by showing that SA aggregates similar inputs $\mathbf{x}_i$ into representations $Z$ with cluster structures. Self-attention updates the token features following an IB principle, where the key matrix $K$ stores the temporary cluster center features  $\mu_c$ and the input features $\mathbf{x}$ are clustered to them via soft association (softmax). The new cluster center features $\mathbf{z}$ are output as the updated token features. The stacked SA modules in ViTs can be broadly regarded as an iterative repeat of this optimization which promotes grouping and noise filtering.

\textbf{Multi-head Self-attention (MHSA).} Many current Vision Transformer architectures adopt an MHSA design where each head tends to focus on different object components. In some sense, MHSA can be interpreted as a mixture of information bottlenecks. We are interested in the relation between the number of heads versus the robustness under a fixed total number of channels. As shown in Figure \ref{fig:head_ablation}, having more heads leads to improved expressivity and robustness. But the reduced channel number per head also causes decreased clean accuracy. The best trade-off is achieved with 32 channels per head.

\subsection{Fully Attentional Networks}

With the above mixture of IBs interpretation, we intend to design a channel processing module that strengthens robust representation through the aggregation across different heads. Our design is driven by two main aspects: 1) To promote more compositional representation, it is desirable to introduce channel reweighting since some heads or channels do capture more significant information than the others. 2) The reweighting mechanism should involve more spatially holistic consideration of each channel to leverage the promoted grouping information, instead of making ``very local'' channel aggregation decisions.

\begin{figure}[t] 
\centering
\includegraphics[width=1.0\linewidth]{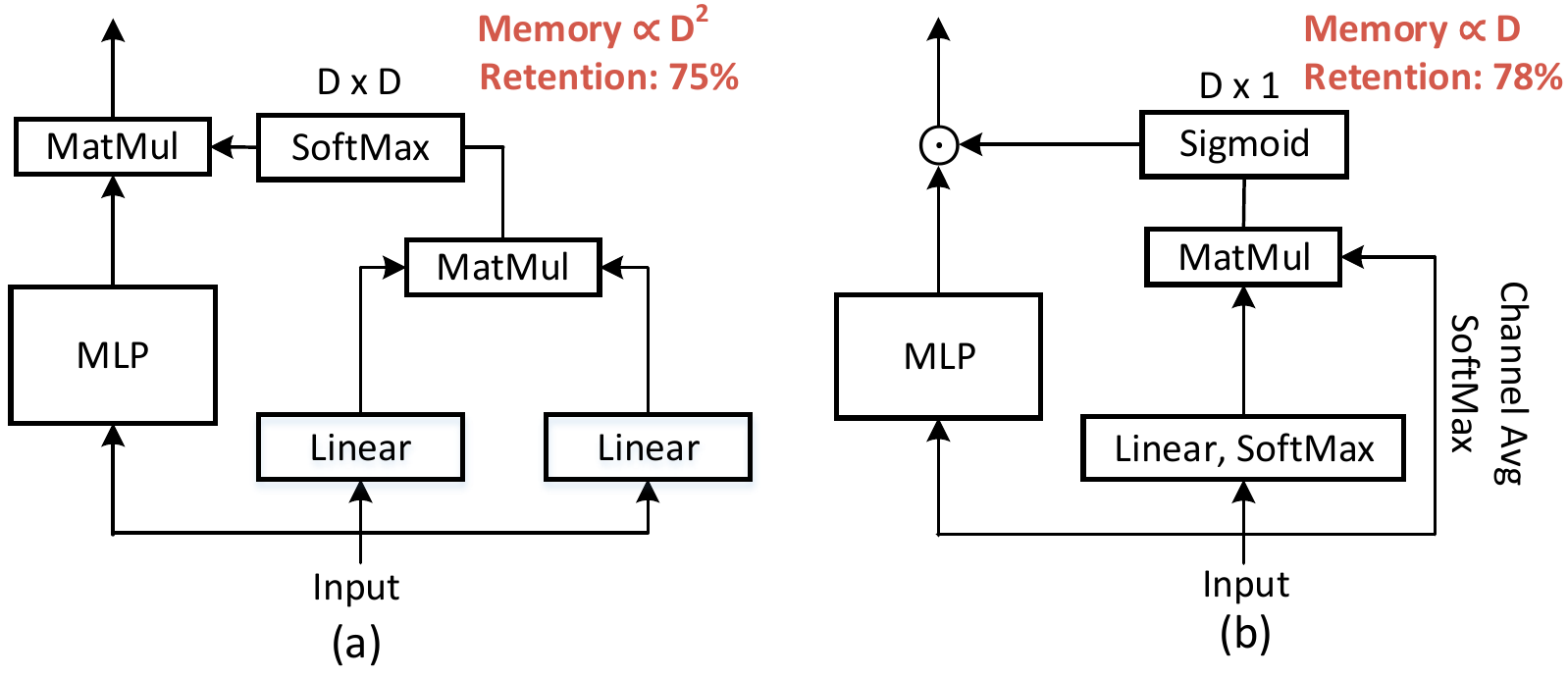}
\caption{\textbf{Comparison among channel attention designs.} (a) CA: a channel self attention design similar to XCiT \cite{el2021xcit}, but differently applied on the output of the MLP block. (b) The proposed efficient channel attention (ECA). 
}
\vspace{-3mm}
\label{fig:csa_arch}
\end{figure}

A starting point towards the above goals is to introduce a channel self-attention design similar to XCiT \cite{el2021xcit}. As shown in Figure~\ref{fig:csa_arch} (a), the channel attention (CA) module adopts a self-attention design which moves the MLP block into the self-attention block, followed by matrix multiplication with the $D\times D$ channel attention matrix from the channel attention branch.

\textbf{Attentional feature transformation.} A FAN block introduces the following channel attention (CA) to perform feature transformation which is formulated as:

\vspace{-0.4cm}
\begin{small}
\begin{equation}
\label{eqn:csa}
    \mathrm{CA}(Z) = \mathrm{Softmax}\left(\frac{(W'_Q Z) (W'_K Z)^\top}{\sqrt{n}}\right)\mathrm{MLP}(Z),
\end{equation}
\end{small}

\vspace{-0.1cm}
Here $W'_Q \in \mathbb{R}^{d\times d}$ and $W'_K \in \mathbb{R}^{d\times d} $ are linear transformation parameters.  
Different from SA, CA  computes the attention matrix along the channel dimension instead of the token  dimension (recall $Z \in \mathbb{R}^{d\times n}$), which leverages the feature covariance (after linear transformation $W'_Q,W'_K$) for   feature  transformation. Strongly correlated feature channels with larger correlation values  will be aggregated while outlier features with low correlation values will be isolated.  This   aids the model in filtering out irrelevant information. 
With the help of CA, the model can filter irrelevant features and thus form more precise token clustering for the foreground and background tokens. We will give a more formal description on such effects in the following section.

We will verify the improved robustness from CA over existing ViT models in the rest of the paper.

\subsection{Efficient Channel Self-attention} There are two limits of applying the conventional self-attention calculation mechanism along the channel dimension. The first one is the \textbf{computational overhead}. 
The computational complexity of CA introduced in Eqn~\ref{eqn:csa} is quadratically proportional to $D^2$, where $D$ is the channel dimension. For modern pyramid model designs \cite{wang2021pyramid, liu2021swin}, the channel dimension becomes larger and larger at the top stages. Consequently, direct applying CA can cause a large computational overhead. 
The second one is the {\bf low parameter efficiency}. In conventional SA module, the attention distribution of the attention weights is sharpened via a Softmax operation. Consequently, only a partial of the channels could contribute to the representation learning as most of the channels are diminished by being multiplied with a small attention weights.
To overcome these, we explore a novel self-attention like mechanism that is equipped with both the high computational efficiency and parameter efficiency.
Specifically, two major modifications are proposed. First, instead of calculating the co-relation matrix between the tokens features, we first generate a token prototype, $\overline{Z}$, $\overline{Z}$ $\in$ $\textit{R}^{n \times 1}$, by averaging over the channel dimension. Intuitively, $\overline{Z}$ aggregates all the channel information for each spatial positions represented by tokens. Thus, it is informative to calculate the co-relation matrix between the token features and token prototype $\overline{Z}$, resulting in learn complexity with respect to the channel dimension.
Secondly, instead of applying a Softmax function, we use a Sigmoid function for normalizing the attention weights and then multiply it with the token features instead of using MatMul to aggregate channel information. Intuitively, we do not force the channel to select only a few of the ``important'' token features but re-weighting each channel based on the spatial co-relation. Indeed, the channel features are typically considered as independent. A channel with large value should not restrain the importance of other channels.
By incorporating those two design concepts, we propose a novel channel self-attention and it is calculated via Eqn. (\ref{eqn:efficient_csa}):

\vspace{-0.4cm}
\begin{small}
\begin{equation}
\label{eqn:efficient_csa}
    \mathrm{ECA}(Z) = \mathrm{Norm}\left(\frac{(W'_Q~ \sigma(Z))~  \sigma(\overline{Z})^\top}{\sqrt{n}}\right)\odot \mathrm{MLP}(Z),
\end{equation}
\end{small}
\vspace{-0.1cm}

Here, $\sigma$ denotes the Softmax operation along the token dimension and $\overline{Z}$ denotes the token prototype ($\overline{Z}$ $\in$ $R^{1 \times N}$).We use sigmoid as the $\mathrm{Norm}$. The detailed block architecture design is also shown in Figure \ref{fig:csa_arch}.  We verify that the novel efficient channel self-attention takes consumes less computational cost while improve the performance significantly. The detailed results will be shown in Sec. \ref{exp:analysis}.

%% file: tex/experiment.tex
\section{Experiment Results \& Analysis}

\subsection{Experiment details}

\textbf{Datasets and evaluation metrics.}  We verify the model robustness on Imagenet-C (IN-C), Cityscape-C and COCO-C without extra corruption related fine-tuning. The suffix `-C' denotes the corrupted images based on the original dataset with the same manner proposed in \cite{hendrycks2019benchmarking}. To test the generalization to other types of out-of-distribution (OOD) scenarios, we also evaluate the accuracy on ImageNet-A \cite{hendrycks2021natural} (IN-A) and ImageNet-R (IN-R) \cite{hendrycks2019benchmarking}. In the experiments, we evaluate the performance with both the clean accuracy on ImageNet-1K (IN-1K) and the robustness accuracy on these out-of-distribution benchmarks. To quantify the resilience of a model against corruptions, we propose to calibrate with the clean accuracy. We use retention rate (Ret R) as the robustness metric, defined as  $R = \frac{\text{Robust Acc.}}{\text{Clean Acc.}} = \frac{\text{IN-C}}{\text{IN-1K}}$. We also report the mean corruption error (mCE) following \cite{hendrycks2019benchmarking}. For more details, please refer to Appendix \ref{app:implementation_details}. For Cityscapes, we take the average mIoU for three severity levels for the noise category, following the practice in SegFormer \cite{xie2021segformer}. For all the rest of the datasets, we take the average of all five severity levels.

\textbf{Model selection.}
We design four different model sizes (Tiny, Small, Base and large) for our FAN models, abbreviated as `-T', `-S',  `-B' and `-L' respectively. Their detailed configurations are shown in Table \ref{tab:model_arch}. For ablation study, we use ResNet-50 as a representative model for CNNs and ViT-S as a representative model for the conventional vision transformers. ResNet-50 and ViT-S have similar model sizes and computation budget as FAN-S. When comparing with SOTA models, we take the most recent vision transformer and CNN models as baselines. 
\vspace{-3mm}
\begin{table}[h]
    \small
    \caption{Details and abbreviations of different FAN variants.}
    \label{tab:model_arch}
    \vspace{2mm}
    \centering
    \setlength{\tabcolsep}{3pt}
    \begin{tabular}{l|cccccc}  
    Model      & \#Blocks  & Channel Dim. &  \#Heads & Param. & FLOPs \\ 
    \midrule
    FAN-T & 12 & 192 & 4 &   7.3M  &   1.4G    \\  
    FAN-S & 12 & 384  & 8 &  28.3M  &   5.3G    \\ 
    FAN-B & 18  & 448 & 8 &  54.0M  &   10.4G    \\ 
    FAN-L & 24  & 480 & 10 &  80.5M  &  15.8G     \\ 
    \end{tabular}
\end{table}

\vspace{-5mm}
\subsection{Analysis}
\label{exp:analysis}
In this section, we present a series of ablation studies to analyze the contribution of self-attention in model robustness. Since multiple advanced training recipes have been recently introduced,  we first investigate their effects in improving model robustness. We then compare ViTs and CNNs with exactly the same training recipes to  exclude factors other than architecture design that might affect model robustness.

\textbf{Effects of advanced training tricks.}
We empirically evaluate how different training recipes could be used to improve the robustness, with the results reported in Table \ref{tab:robustness_impact_vit}. Interestingly, it is observed that widely used tricks such as knowledge distillation (KD) and large dataset pretraining do improve the absolute accuracy. However, they do not significantly reduce the performance degradation when transferred to ImageNet-C. The main improvement comes from the advanced training recipe such as the CutMix and  RandAugmentation adopted in DeiT training recipe. In the following comparison, we use the ViT-S trained with DeiT recipe and increased block number with reduced channel dimension, denoted as ViT-S$^*$. In addition, to make   fair comparison, we first apply those advanced training techniques to reproduce the ResNet-50 performance.
\vspace{-3mm}
\begin{table}[h]
    \small
    \caption{Impacts of various performance improvement tricks on model robustness (\%).}
    \label{tab:robustness_impact_vit}
    \vspace{2mm}
    \centering
    \setlength{\tabcolsep}{4pt}
    \begin{tabular}{l|cccc}  %\toprule
    Model      & IN-1K      & IN-C   &  Retention &mCE ($\downarrow$) \\
    \midrule
    ViT-S &  77.9 & 54.2 &  70      &  63.5 \\  
    \quad + DeiT Recipe &  79.3  & 57.1 &   72    & 57.1 \\ 
    \quad +  \#Blocks (8 $\uparrow$12) & 79.9  & 58.0 & 72 &  56.2 \\
    \quad + KD & 81.3 & 59.6 &      73 & 54.0 \\ 
    \quad + IN22K w/o KD &  81.8 & 59.7 &  73 & 54.2 \\
    \end{tabular}
\end{table}

\vspace{-5mm}
\textbf{Adding new training recipes to CNNs.}
We make a step by step empirical study on how the robustness of ResNet-50 model changes when adding advanced tricks. We examine three design choices: training recipe, attention mechanism and down-sampling methods.
For the training recipe, we adopt the same one as used in training the above ViT-S model. We use Squeeze-and-Excite (SE) attention \cite{hu2018squeeze} and apply it along the channel dimension for the feature output of each block. We also investigate different downsampling strategies, i.e., average pooling (ResNet-50 default) and strided convolution. The results are reported in Table \ref{tab:cnn_impact}. As can be seen, adding attention (Squeeze-and-Excite (SE) attention) and using more advanced training recipe do improve the robustness of ResNet-50 significantly. We take the best-performing ResNet-50 with all these tricks, denoted as ResNet-50$^*$, for the following comparison.

\begin{table}[h]
    \small
    \caption{Robustness of ResNet-50 with various performance improvement tricks (\%).}
    \label{tab:cnn_impact}
    \vspace{2mm}
    \setlength{\tabcolsep}{2pt}
    \centering
    \begin{tabular}{l|cccc}  
    Model &IN-1K   &IN-C    & Retention& mCE ($\downarrow$)  
    \\ 
    \midrule 
    ResNet-50 &  76.0 & 38.8 &  51  &  76.7   
    \\  
    \quad + DeiT Recipe &  79.0  & 43.9 &     46 & 69.7     
    \\ 
    \quad + SE &  79.8  & 50.1 &     63 & 63.1    
    \\ 
    \quad + Strided Conv &  80.2  & 52.1 &     65 & 61.6     
    \vspace{-3mm}
    \end{tabular}
\end{table}

\textbf{Advantages of ViTs over CNNs on robustness.} To make fair comparison, we use all the above validated training tricks to train the ViT-S and ResNet-50 to their best performance. Specifically, ResNet-50$^*$ is trained with DeiT recipe, SE and strided convolution; ViT-S$^*$ is also trained with DeiT recipe and has 12 blocks with 384 embedding dimension for matching the model size as ResNet-50. Results in Table \ref{tab:robustness_compare_vit_cnn} show that even with the same training recipe, ViTs still outperform ResNet-50 in robustness. These results indicate that the improved robustness in ViTs may come from their architectural advantages with self-attention. This motivates us to further improve the architecture of ViTs by leveraging self-attention more broadly to further strengthen the model's robustness.
\vspace{-4mm}
\begin{table}[h]
    \small
    \caption{Robustness comparison between ResNet-50 and ViT-S (\%).}
    \label{tab:robustness_compare_vit_cnn}
    \vspace{2mm}
    \centering
    \setlength{\tabcolsep}{4pt}
    \begin{tabular}{l|c|cccc}  
    % \toprule
    Model      & Param & IN-1K     & IN-C  &  Retention & mCE ($\downarrow$) \\
    \midrule
    ResNet-50$^*$& 25M &  80.2  & 52.1 &  65  & 61.6       \\  
    ViT-S$^*$& 22M  & 79.9    & 58.0 &  72  & 56.2      \\ 
    
    \end{tabular}
\end{table}

\vspace{-3mm}
\textbf{Difference among ViT, Swin-ViT and ConvNeXt.} Some latest CNN architectures~\cite{liu2022convnet} have also shown superior robustness over Swin Transformer. We interpret this phenomenon from the following perspective: SA module uses softmax to normalize each row of the attention matrix. Such softmax-based normalization encourages the competition among the attention to different tokens, thus promoting the selection of particular tokens. As Swin Transformer adopts window based local self-attention, the selection is constrained within a predefined window area. When a window contains no essential information, this local design may also promote the selection of some unimportant tokens, resulting in spurious and suboptimal representations. 
% The robustness of Swin Transformer is thus influenced.
However, as shown in Table \ref{tab:comp_conv_vit_fan}, DeiT achieves better robustness with 24.1\% less number of parameters compared to ConvNeXt. This indicates that Transformers with a global SA design still maintains a competitive edge over state-of-the-art CNN models in terms of robustness.

% Latest CNN architecture  has shown superiority of the robustness over the recent state-of-the-art transformer based models Swin transformer. We here interpret this from the view of information bottleneck. As explained in Sec. \ref{sec:explain}, the SA module is forming an IB to select essential tokens. As Swin transformer deploys a window based local self-attention mechanism, it forces the model to select information from a predefined window area. Such a local window IB forces each window to select tokens from a local constrained features. Intuitively, when a selected window contains no essential information, a local SA is forced to select some key tokens and thus resulting a set of sub-optimal clusters. Thus, the robustness of Swin transformer is worse than the recent SOTA CNN model ConvNeXt. However, as shown in Table \ref{tab:comp_conv_vit_fan}, DeiT achieve better robustness with 24.1\% less number of parameters, compared to ConvNeXt model. We thus argue that transformers with global SA module are still more robust than the state-of-the-art ConvNeXt model.

\begin{table}[h!]
\centering
\small
\caption{\textbf{Robustness comparison among Swin, ConvNeXt, DeiT and FAN}. The mIoU of ConvNeXt, DeiT, Swin and SegFormer models are our reproduced results. 
}
\label{tab:comp_conv_vit_fan}
\vspace{2mm}
\setlength{\tabcolsep}{2pt}
\resizebox{\linewidth}{!}{
\begin{tabular}{l|c|ccc|ccc}
\multirow{2}{*}{Model} & \multirow{2}{*}{Param.} & \multicolumn{3}{c|}{ImageNet} & \multicolumn{3}{c}{Cityscapes}  \\
\cline{3-8}
&  & Clean & Corrupt & Reten. & Clean &  Corrupt & Reten.  \\
\midrule
\multicolumn{1}{l|}{ConvNeXt~(\citeauthor{liu2022convnet})} & 29M & {82.1} & 59.1 & 72.0  & {79.0} & 54.2 & 68.6   \\
\multicolumn{1}{l|}{SWIN~(\citeauthor{liu2021swin})} & {28M} & 81.3 & 55.4 & 68.1 & 78.0 & 47.3 & 61.7  \\
\multicolumn{1}{l|}{DeiT-S~(\citeauthor{touvron2021training})} & {22M} & 79.9 & 58.1 & 72.7 & {76.0} & 55.4 & 72.9    \\
\multicolumn{1}{l|}{FAN-Hybrid-S (Ours)} & {26M} & 83.5 & 64.7 & 78.2 & {81.5} & 66.4 & 81.5   \\
\end{tabular}
}
\vspace{-2mm}
\end{table}

\subsection{Fully Attentional Networks}
In this subsection, we investigate how the new FAN architecture improves the model's robustness among different architectures. 

\textbf{Impacts of efficient channel attention} We first ablate the impacts of different forms of channel attentions in terms of GPU memory consumption, clean image accuracy and robustness. The results are shown in Table \ref{tab:attn_ablate}. Compared to the original self-attention module, SE attention consumes less memory and achieve comparable clean image accuracy and model robustness. By taking the spatial relationship into consideration, our proposed CSA produces the best model robustness with comparable memory consumption to the SE attention.

\vspace{-4mm}
\begin{table}[h]
    \small
    \caption{Effects of different channel attentions on model robustness (\%).}
    \label{tab:attn_ablate}
    \vspace{2mm}
    \centering
    \setlength{\tabcolsep}{2.5pt}
    \resizebox{\linewidth}{!}{
    \begin{tabular}{l|ccccc}  
    Model   & Mem.(M)   & IN-1K  & IN-C    &  Retention   & mCE ($\downarrow$) 
    \\ \midrule
    FAN-ViT-S-SA & 235 & 81.3  & 61.7&  76  & 51.4      
    \\  
    FAN-ViT-S-SE & 126 & 81.2  & 62.0 &  76  & 50.0     
    \\  
    FAN-ViT-S-ECA & 127 & 82.5  & 64.6 &  78  & 47.7 
    \\  
    \end{tabular}
    }
\end{table}

% \vspace{-1mm}
\textbf{FAN-ViT \& FAN-Swin.}
 Using the  FAN block  to replace the conventional transformer block forms the FAN-ViT. FAN-ViT  significantly enhances the robustness. 
However,   compared to ViT,  the robustness of Swin architecture \cite{liu2021swin} (which uses shifted window attention)  drops.  This is possibly because their local attention hinders global clustering  and  the IB-based   information extraction, as detailed in Section \ref{exp:analysis}.  The drop in robustness can be effectively remedied by using the FAN block. By adding the ECA to the feature transformation of SWIN models, we obtain the FAN-SWIN, a new FAN model whose spatial self-attention is augmented by the shifted window attention in SWIN.   As shown in Table \ref{tab:robustness_vit_arch},   adding FAN block improves the accuracy on ImageNet-C by 5\%. Such a significant improvement shows that our proposed CSA does have significant effectiveness on improving the model robustness. 

\vspace{-4mm}
\begin{table}[h]
    \small
    \caption{Effects of architectural changes on model robustness (\%).}
    \label{tab:robustness_vit_arch}
    \vspace{2mm}
    \centering
    \setlength{\tabcolsep}{4pt}
    \begin{tabular}{l|cccc}  
    Model      & IN-1K       & IN-C    &  Retention   & mCE ($\downarrow$)  
    \\
    \midrule
    ViT-S$^*$ & 79.9  & 58.1 &  73  & 56.2      
    \\  
    \quad + FAN  &  81.3  & 61.7 &  76  &  51.4
     \\ 
    \midrule
     Swin-T &81.4 &55.4 &68 &  59.6\\
    \quad + FAN  &81.9 &59.4 &73 & 54.5 \\
    \midrule
    ConvNeXt-T &82.1 &59.1 &72 & 54.8 \\
    \quad + FAN & 82.5 & 60.8 & 74 & 53.1
    \end{tabular}
\end{table}

\vspace{-1mm}
\textbf{FAN-Hybrid.}
From the clustering process as presented in Figure \ref{fig:clustering_number}, we find that the clustering mainly emerges   at the top stages of the FAN model, implying the bottom stages to focus on extracting  local visual patterns. Motivated by this,  we  propose to use convolution blocks for the bottom two stages with  down-sampling and then append FAN blocks to the output of the convolutional stages. Each stage includes 3 convolutional blocks. This gives the FAN-Hybrid model. In particular, we use the ConvNeXt \cite{liu2022convnet},  a very recent  CNN model, to build the early stages of our hybrid model. As shown in Table~\ref{tab:robustness_vit_arch}, we find original ConvNeXt exhibits strong robustness than SWIN transformer, but performs less robust than FAN-ViT and FAN-Swin models. However, the FAN-Hybrid achieves comparable robustness as FAN-ViT and FAN-SWIN and presents higher accuracy for both clean and corrupted datasets, implying FAN can also effectively strengthen the robustness of a CNN-based model. Similar to FAN-SWIN, FAN-Hybrid enjoys  efficiency for processing large-resolution inputs and dense prediction tasks, making it favorable    for downstream tasks. Thus, for all downstream tasks, we use FAN-Hybrid model to compare with other state-of-the-art models. More details on the FAN-Hybrid and FAN-SWIN architecture can be found in the appendix. 

\subsection{Comparison to SOTAs on various tasks}
In this subsection, we evaluate the robustness of FAN with other SOTA methods against common corruptions on different downstream tasks, including image classification (ImageNet-C), semantic segmentation (Cityscapes-C) and object detection (COCO-C). Additionally, we evaluate the robustness of FAN on various other robustness benchmarks including ImageNet-A and ImageNet-R to further show its non-trivial improvements in robustness.

\textbf{Robustness in image classification.}
We first compare the robustness of FAN with other SOTA models   by directly applying them (pre-trained on ImageNet-1K) to the ImageNet-C dataset \cite{hendrycks2019benchmarking} without any fine-tuning. We divide all the models into three groups according to their model size for fair comparison.  The results are shown in Table \ref{tab:robustness_imagenet_c} and the detailed results are summarized in Table \ref{tab:robustness_imc}. From the results, one can clearly observe that all the transformer-based models show stronger robustness than   CNN-based models. Under all the models sizes, our proposed FAN models surpass all other models significantly. They offer strong robustness to all the types of corruptions. Notably, FANs perform excellently robust for bad weather conditions and digital noises, making   them very suitable for vision applications in mobile phones and self-driving cars. 

\begin{table}[h]
    \small
    \caption{\textbf{Main results on image classification.} FAN models show improved performance in both clean accuracy and robustness than other models. $\dagger$ denotes models are pretrained on ImageNet-22K.}
    \label{tab:robustness_imagenet_c}
    \vspace{2mm}
    \centering
    \setlength{\tabcolsep}{3pt}
    \resizebox{\linewidth}{!}{
    \begin{tabular}{l|cccc}  
    Model  &  Param./FLOPs   & IN-1K       & IN-C    &  Retention  
    \\
    \midrule
    ResNet18 (\citeauthor{he2016deep}) & 11M/1.8G & 69.9  & 32.7 & 46.8\%     
    \\  
    MBV2 (\citeauthor{sandler2018mobilenetv2})  &  4M/0.4G  & 73.0 &  35.0  & 47.9\% 
     \\ 
    EffiNet-B0 (\citeauthor{tan2019efficientnet})  &  5M/0.4G  & 77.5 &  41.1  & 53.0\%  
     \\ 
    PVTV2-B0 (\citeauthor{wang2021pyramid})  &  3M/0.6G  & 70.5 &  36.2  & 51.3\% 
     \\ 
    PVTV2-B1 (\citeauthor{wang2021pyramid})  &  13M/2.1G  & 78.7 &  51.7  &  65.7\% 
     \\ 
     \midrule
         FAN-T-ViT  &  7M/1.3G  & 79.2 &  57.5  & \textbf{72.6\%} 
     \\ 
         FAN-T-Hybrid  &  7M/3.5G  & 80.1 &  57.4  & 71.7\% 
     \\ 
     \midrule
    ResNet50 (\citeauthor{he2016deep})  &  25M/4.1G  & 79 &  50.6  & 64.1\% 
     \\ 
         DeiT-S (\citeauthor{touvron2021training})  &  22M/4.6G  & 79.9 &  58.1  & 72.7\% 
     \\ 
         Swin-T (\citeauthor{liu2021swin})  &  28M/4.5G  & 81.3 &  55.4  & 68.1\% 
     \\ 
         ConvNeXt-T (\citeauthor{liu2022convnet})  &  29M/4.5G  & 82.1 &  59.1  & 71.9\% 
     \\ 
     \midrule
         FAN-S-ViT  &  28M/5.3G  & 82.9 &  64.5  & \textbf{77.8\%}
     \\ 
              FAN-S-Hybrid  &  26M/6.7G  & 83.5 &  64.7  & 77.5\% 
     \\ 
    \midrule
         Swin-S(\citeauthor{liu2021swin})  &  50M/8.7G  & 83.0 &  60.4  & 72.8\% 
     \\ 
         ConvNeXt-S (\citeauthor{liu2022convnet})  &  50M/8.7G  & 83.1 & 61.7   & 74.2\% 
     \\ 
     \midrule
         FAN-B-ViT  &  54M/10.4G  & 83.6 &  67.0  & 80.1\% 
     \\ 
         FAN-B-Hybrid  &  50M/11.3G  & 83.9 &  66.4  & 79.1\% 
     \\ 
         FAN-B-Hybrid$^\dagger$  &  50M/11.3G  & 85.6 &  70.5  & \textbf{82.4\%} 
     \\ 
    \midrule
         DeiT-B (\citeauthor{touvron2021training})  &  89M/17.6G  & 81.8 &  62.7  &  76.7\% 
     \\ 
         Swin-B (\citeauthor{liu2021swin})  &  88M/15.4G  & 83.5 &  60.4  & 72.3\% 
     \\ 
         ConvNeXt-B (\citeauthor{liu2022convnet})  &  89M/15.4G & 83.8  & 61.7 &  73.6\%
     \\ 
     \midrule
         FAN-L-ViT  &  81M/15.8G  & 83.9 &  67.7  & 80.7\% 
     \\ 
         FAN-L-Hybrid  &  77M/16.9G  & 84.3 &  68.3  & {81.0\%} 
     \\ 
         FAN-L-Hybrid$^\dagger$  &  77M/16.9G  & 86.5 &  73.6  & \textbf{85.1\%} 
     \\ 
    \bottomrule
    \end{tabular}
    }
\vspace{-6mm}
\end{table}

We also evaluate the zero-shot robustness of the Swin transformer and the recent ConvNeXt. Both of them demonstrate weaker robustness than the transformers with global self-attention. However, adding FAN to them improves their robustness, enabling  the resulted FAN-SWIN and FAN-Hybrid variants to inherit both high  applicability for downstream tasks  and strong robustness to corruptions. We will use FAN-Hybrid variants in the applications of segmentation and detection.

\begin{figure*}[t] 
\centering
\includegraphics[width=1\linewidth]{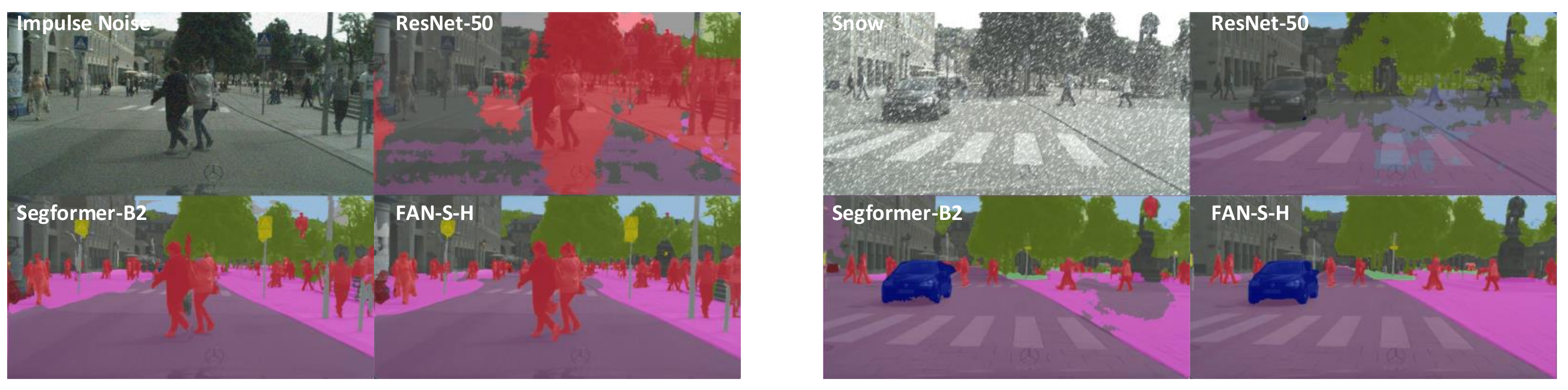}
\vspace{-5mm}
\caption{\textbf{Segmentation visualization on corrupted images with impulse noise (severity 3) and snow (severity 3).} We select the recent state-of-the-art SegFormer model \cite{xie2021segformer} as a strong baseline. FAN-S-H denotes our hybrid model. Under comparable model size and computation, FAN achieve significantly improved segmentation results over ResNet-50 and SegFormer-B2 model. A video demo is available via \href{https://drive.google.com/file/d/1A9Roe7SanJV145qh481nDpf5JAmiJ-Fl/view?usp=sharing}{external players}.}

\label{fig:segmentation_vis}
\vspace{-5mm}
\end{figure*}

\textbf{Robustness in semantic segmentation.} We further evaluate robustness of our proposed FAN model for the segmentation task. We use the Cityscapes-C for evaluation, which expands the Cityscapes validation set with 16 types of natural corruptions. We compare our model to variants of DeeplabV3+ and latest SOTA models.  The results are summarized in Table \ref{tab:robustness_cityscape_c} and by category results are summarized in Table \ref{tab:benchmark_cityscapes}. Our model significantly outperforms previous models. FAN-S-Hybrid surpasses the latest SegFormer\textemdash a transformer based segmentation model\textemdash by 6.8\% mIoU with comparable model size.  The results indicate strong robustness of FAN. 

\begin{table}[h]
    \small
    \caption{\textbf{Main results on semantic segmentation.} `R-' and `X-' refer to DeepLabv3+, ResNet and Xception. The mIoUs of DeepLabv3+ framework  are reported from \cite{kamann2020benchmarking}. FAN shows significantly stronger clean accuracy and robustness than other models.}
    \label{tab:robustness_cityscape_c}
    \vspace{2mm}
    \centering
    \setlength{\tabcolsep}{3pt}
    \resizebox{\linewidth}{!}{
    \begin{tabular}{l|cccc}
    Model  &  Encoder Size   & City       & City-C    &  Retention 
    \\
    \midrule
    DeepLabv3+ (R50)  &  25.4M  & 76.6 &  36.8  & 48.0\% \\ 
    DeepLabv3+ (R101)  &  47.9M  & 77.1 &  39.4  &  51.1\% \\ 
    DeepLabv3+ (X65)  & 22.8M  & 78.4 &  42.7  &  54.5\% \\ 
    DeepLabv3+ (X71)  & -  & 78.6 &  42.5  &  54.1\% \\ 
    \midrule
    ICNet (\citeauthor{zhao2018icnet})  &  -  & 65.9 &  28.0  & 42.5\% \\ 
    FCN8s (\citeauthor{long2015fully})  &  50.1M  & 66.7 &  27.4  & 41.1\% \\ 
    DilatedNet (\citeauthor{yu2015multi})  &  -  & 68.6 &  30.3  &  44.2\% \\ 
    ResNet38 (\citeauthor{wu2019wider})  &  -  & 77.5 &  32.6  &  42.1\% \\ 
    PSPNet (\citeauthor{zhao2017pyramid})  &  13.7M  & 78.8 &  34.5  &  43.8\% \\ 
    ConvNeXt-T (\citeauthor{liu2022convnet})  &  29.0M  & 79.0 &  54.4  &  68.9\% \\ 
    \midrule
    SETR (\citeauthor{heo2021rethinking}) &  22.1M  & 76.0 &  55.3  & 72.8\% \\ 
    Swin-T (\citeauthor{liu2021swin})  &  28.4M  & 78.1 &  47.3  &  60.6\% \\ 
    SegFormer-B0 (\citeauthor{xie2021segformer})  &  3.4M  & 76.2 &  48.8  &  64.0\% \\ 
    SegFormer-B1 (\citeauthor{xie2021segformer})  & 13.1M  & 78.4 &  52.7  &  67.2\% \\
    SegFormer-B2 (\citeauthor{xie2021segformer})  &  24.2M  & 81.0 &  59.6  &  73.6\% \\ 
    SegFormer-B5 (\citeauthor{xie2021segformer})  &  81.4M  & 82.4 &  65.8  &  79.9\% \\ 
    \midrule
    FAN-T-Hybrid (Ours)  &  7.4M  & 81.2 &  57.1  &  70.3\%  \\ 
    FAN-S-Hybrid (Ours)   & 26.3M  & 81.5 &  66.4  &  81.5\% \\
    FAN-B-Hybrid (Ours)   &  50.4M  & 82.2 &  66.9  &  81.5\% \\ 
    FAN-L-Hybrid (Ours)   & 76.8M &  82.3  & 68.7   & \textbf{83.5\%} \\ 
    \bottomrule
    \end{tabular}
    }
\vspace{-6mm}
\end{table}

\textbf{Robustness in object detection.} We also evaluate the robustness of FAN  for the detection task on COCO-C dataset, an extension of COCO generated similarly as Cityscapes-C. The results are summarized in Table \ref{tab:robustness_coco_c} and the detailed results are summarized in Table \ref{tab:robustness_coco}. FAN demonstrates strong robustness again, yielding improvement over recent SOTA Swin transformer \cite{liu2021swin} by 6.2\% mAP with comparable model size (26M vs 29M) under same training settings and showing a new state-of-the-art results of 42.0\% mAP with only 76.8M number of parameters for the encoder model. 

\begin{table}[h]
    \small
    \caption{\textbf{Main results on object detection.} FAN shows stronger clean accuracy and robustness than other models. `$\dagger$' denotes the accuracy pretrained on ImageNet-22K.}
    \label{tab:robustness_coco_c}
    \vspace{1mm}
    \centering
    \setlength{\tabcolsep}{4pt}
    \resizebox{\linewidth}{!}{
    \begin{tabular}{l|cccc}  
    Model & Encoder Size & COCO & COCO-C & Retention \\
    \midrule
    \multicolumn{5}{c}{Mask R-CNN} \\
    \midrule
    ResNet-50 \cite{he2016deep}  & 25.4M  & 39.9 &  21.3  & 53.3\% \\  
    ResNet101 (\cite{he2016deep})  &  44.1M  & 41.8 &  23.3  &  55.7\% \\ 
    DeiT-S \cite{touvron2021training} &  22.1M  & 40.0 &  26.9  & 67.3\% \\ 
    Swin-T \cite{liu2021swin}  &  28.0M & 46.0 &  29.3  &  63.7\% \\ 
    FAN-T-Hybrid & 7.4M  & 45.8 &  29.7  &  64.8\% \\ 
    FAN-S-Hybrid & 26.3M  & 49.1 &  35.5  &  72.3\% \\
    \midrule
    \multicolumn{5}{c}{Cascade Mask R-CNN} \\
    \midrule
    FAN-T-Hybrid & 7.4M  & 50.2 &  33.1  &  65.9\% \\ 
    FAN-S-Hybrid & 26.3M  & 53.3 &  38.7  &  72.6\% \\
    FAN-L-Hybrid   & 76.8M &  54.1  & 40.6   & {75.0\%} \\
    FAN-L-Hybrid$^\dagger$    & 76.8M &  55.1  & 42.0   & \textbf{76.2\%} \\ 
    \bottomrule
    \end{tabular}
    }
\vspace{-4mm}
\end{table}

\textbf{Robustness against out-of-distribution.} The FAN encourages token features to form clusters and implicitly selects the informative features, which would benefit   generalization performance of the model. To verify this, we directly test our ImageNet-1K trained models for evaluating their robustness, in particular for  out-of-distribution samples,  on ImageNet-A and ImageNet-R. The experiment results are summarized in Table \ref{tab:generalization}. Among these models, ResNet-50 (\citeauthor{liu2021swin}) presents weakest generalization ability while the recent ConvNeXt substantially improves the generalization performance of CNNs. The transformer-based models, Swin and RVT performs comparably well as ConvNeXt and much better than ResNet-50. Our proposed FANs outperform  all these models significantly, implying the fully-attentional architecture aids generalization ability of the learned representations as the irrelevant features are effectively processed.

\input{tables/cross_comparison_multi_dataset}

%% file: tables/cross_comparison_multi_dataset.tex
\begin{table}[t]
\vspace{-2mm}
% \tablestyle{7.2pt}{1.0}
\caption[caption]{\textbf{Main results on out-of-distribution generalization.} FAN models show improved generalization across all datasets. `$^\ddagger$' denotes results with finetuning on 384 $\times$ 384 image resolution. IN-C is measured by mCE ($\downarrow$). All metrics are scaled by (\%).}
\vspace{2mm}
\setlength{\tabcolsep}{1pt} 
\resizebox{\linewidth}{!}{
\begin{tabular}{l|cccccc  }
        Model  &  Params (M) 
        & Clean 
        &  ~IN-A~  \quad & ~IN-R~ \quad& ~IN-C~ \\
        \midrule
        &\quad ImgNet-1k & pretrain  \\
        \midrule
        XCiT-S12~(\citeauthor{el2021xcit})~ &   26.3 
        & 81.9 
        & 25.0 & 45.5 & 51.5  \\
        XCiT-S24~(\citeauthor{el2021xcit})~  &   47.7
        & 82.6
        & 27.8 & 45.5 & 49.4    \\
        \midrule
        RVT-S*~(\citeauthor{mao2021towards}) &   23.3 
        & 81.9 
        & 25.7 & 47.7 & 51.4  \\
        RVT-B*~(\citeauthor{mao2021towards})  &   91.8 
        & 82.6 
        & 28.5 & 48.7 & 46.8    \\
        \midrule
        Swin-T~(\citeauthor{liu2021swin})  &   28.3 
        & 81.2 
        & 21.6 & 41.3 & 59.6 \\
        Swin-S~(\citeauthor{liu2021swin})   &   50
        &  {83.4} 
        & 35.8 & 46.6 & 52.7 \\
        Swin-B~(\citeauthor{liu2021swin})   &  87.8
        &  {83.4} 
        & 35.8 & 64.2 & 54.4 \\
        \midrule
        {ConvNeXt-T} (\citeauthor{liu2022convnet}) &    28.6 
        & 82.1 
        & 24.2 & 47.2 & 53.2  \\
        {ConvNeXt-S} (\citeauthor{liu2022convnet}) &    50.2 
        & 82.1 
        & 31.2 & 49.5 & 51.2  \\
        {ConvNeXt-B} (\citeauthor{liu2022convnet}) &    88.6 
        & 83.8 
        & 36.7 & 51.3 & 46.8  \\
        \midrule
        MAE-ViT-B~(\citeauthor{he2022masked})   &  86
        &  {83.6} 
        & 35.9 & 48.3 & 51.7 \\
        MAE-ViT-L~(\citeauthor{he2022masked})   &  307
        &  {85.9} 
        & 57.1 & 59.9 & 41.8 \\
        \midrule
        {FAN-S-ViT}  (Ours) &  28.0 
        & 82.5 
        & 29.1 &  50.4 & {47.7} \\
        {FAN-B-ViT}  (Ours) &  54.0  
        & 83.6 
        & {35.4}  &  {51.8} & {44.4} \\
        {FAN-L-ViT  (Ours)}  &   80.5  
        & {83.9} 
        & {37.2} &  {53.1} &  {43.3}  \\
        \midrule
        {FAN-S-Hybrid}  (Ours) &  26.0 
        & 83.6
        & 33.9 &  50.7 & {47.8} \\
        {FAN-B-Hybrid}  (Ours) &  50.0  
        & 83.9
        & {39.6}  &  {52.9} & {45.2} \\
        {FAN-L-Hybrid  (Ours)}  &   76.8  
        & {84.3} 
        & {41.8} &  {53.2} &  {43.0}  \\
        \midrule
        & \quad  ImgNet-22k & ~pretrain  \\
        \midrule
        {ConvNeXt-B$^\ddagger$} (\citeauthor{liu2022convnet}) &  88.6  
        & 86.8 
        & {62.3}  &  {64.9} & {43.1} \\
        \midrule
        {FAN-L-Hybrid  (Ours)}  &   76.8  
        & {86.5} 
        & {60.7} &  {64.3} &  \textbf{35.8}  \\
        {FAN-L-Hybrid$^\ddagger$  (Ours)}  &   76.8  
        & \textbf{87.1} 
        & \textbf{74.5} &  \textbf{71.1} &  {36.0}  \\
    \end{tabular}
} 
\label{tab:generalization}
\vspace{-4mm}
\end{table}

%% file: tex/related.tex
\section{Related Works}

Vision Transformers~\cite{vaswani2017attention} are a family of transformer-based architectures  on computer vision tasks.
Unlike CNNs relying on certain inductive biases (e.g., locality and  translation invariance), ViTs perform the global interactions among visual tokens  via self-attention, thus having less inductive bias about the input image data. Such designs have offered significant performance improvement on various vision tasks including image classification~\cite{dosovitskiy2020image,yuan2021tokens,zhou2021deepvit, zhou2021refiner}, object detection~\cite{carion2020end,zhu2020deformable,dai2020up,zheng2020end} and segmentation~\cite{wang2020end,liu2021swin, zheng2020end}. The success of vision transformers for vision tasks triggers
broad debates and studies on the advantages of self-attention
versus convolutions~\cite{raghu2021vision,tang2021sparse}. Compared to convolutions, an important advantage is the robustness against observable corruptions. Several works~\cite{bai2021transformers,xie2021segformer,paul2022vision,naseer2021intriguing} have empirically shown that the robustness of ViTs against corruption  consistently outperforms ConvNets by significant margins.
However, how the key component (i.e. self-attention) contributes to the robustness is under-explored.  In contrast, our work conducts empirical studies to reveal
intriguing properties (i.e., token grouping
and noise absorbing) of self-attention for robustness and presents a novel fully attentional architecture design to further improve the robustness. 

There exists a large body of work on improving robustness of deep learning models in the context of adversarial examples by developing robust training algorithms \cite{kurakin2016adversarial,shao2021adversarial},  which differs from the scope of our work. In this work, we focus the zero-shot robustness to the natural corruptions and mainly study improving model's robustness from the  model architecture   perspective. 

%% file: tex/appendix.tex
%%%%%%%%%%%%%%%%%%%%%%%%%%%%%%%%%%%%%%%%%%%%%%%%%%%%%%%%%%%%%%%%%%%%%%%%%%%%%%%
%%%%%%%%%%%%%%%%%%%%%%%%%%%%%%%%%%%%%%%%%%%%%%%%%%%%%%%%%%%%%%%%%%%%%%%%%%%%%%%
% APPENDIX
%%%%%%%%%%%%%%%%%%%%%%%%%%%%%%%%%%%%%%%%%%%%%%%%%%%%%%%%%%%%%%%%%%%%%%%%%%%%%%%
%%%%%%%%%%%%%%%%%%%%%%%%%%%%%%%%%%%%%%%%%%%%%%%%%%%%%%%%%%%%%%%%%%%%%%%%%%%%%%%
\newpage
\appendix
\onecolumn
\section{Supplementary Details}
\subsection{Proof on the relationship between the Information Bottleneck  and Self-Attention }
Given a  distribution $X \sim \mathcal{N}(X',\epsilon)$ with $X$ being the observed noisy input and $X'$ the target clean code, IB seeks a mapping $f(Z|X)$ such that $Z$ contains the relevant information in $X$ for predicting $X'$.  This goal is formulated as the following information-theoretic optimization problem
\begin{equation}
    q^*_{\mathrm{IB}}(z|x) = \arg\min_{q(z|x)} I(X,Z) - \beta I(Z,X'), 
\end{equation}
subject to the Markov constraint $Z \leftrightarrow X \leftrightarrow X'$. $\beta$ is a free parameter that trades-off the information compression by the first term and the relevant information maintaining by the second. 

The information bottleneck approach can be applied for solving unsupervised clustering problems.  Here we choose $X$ to be the data point with index $i$ that will  be clustered into clusters with  indices $c$.

As mentioned above, we assume the following data distribution:
\begin{equation}
    p(\mathbf{x}|i) \propto \exp \left[-\frac{1}{2\epsilon^2} \|\mathbf{x} - \mathbf{x}_i\|^2 \right],
\end{equation}
where $s$ is a smoothing parameter. We assume the marginal to be $p(i)=\frac{1}{N}$, where $N$ is the number of data points. 

Using the above notations, the $t$-th step in the iterative IB for clustering is formulated as
\begin{equation}
    \begin{aligned}
        q^{(t)}(c|i) &= \frac{\log q^{(t-1)}(c)}{K(\mathbf{x},\beta)}\exp\left[-\beta~\mathrm{KL}[p(\mathbf{x}|i) | q^{(t-1)}(\mathbf{x}|c)]\right], \\
        q^{(t)}(c) &= \frac{n_c^{(t)}}{N}, \\
        q^{(t)}(\mathbf{x}|c) &= \frac{1}{n_c^{(t)}}\sum_{i \in S_c^{(t)}} p(\mathbf{x}|i). 
    \end{aligned}
\end{equation}
Here $K(\mathbf{x},\beta)$ is the normalizing factor and $S_c$ denotes the set of indices of data points assigned to cluster $c$.

We choose to replace $q(\mathbf{x}|c)$ with a Gaussian approximation $g(\mathbf{x}|c)=\mathcal{N}(\mathbf{x}|\mu_c,\Sigma_c)$ and assume $\epsilon$ is sufficiently small. Then,
\begin{equation}
    \mathrm{KL}[  p(\mathbf{x}|i) | g (\mathbf{x}|c)  ] \propto   (\mu_c - \mathbf{x}_i)^\top \Sigma_c^{-1} (\mu_c - \mathbf{x}_i) + \log \det \Sigma_c + B,
\end{equation}
where $B$ denotes terms not dependent on the assignment of data points to clusters and thus irrelevant for the objective. Thus the above cluster update can be written as:
\begin{equation}
\begin{aligned}
    q^{(t)}(c|i) &= \frac{\log q^{(t-1)}(c)}{\det \Sigma_c} \frac{\exp\left[-(\mu_c - \mathbf{x}_i)^\top \Sigma_c^{-1} (\mu_c - \mathbf{x}_i)\right]}{Z(\mathbf{x},\beta)} \\
    & = \frac{\log q^{(t-1)}(c)}{\det \Sigma_c} \frac{\exp\left[-(\mu_c - \mathbf{x}_i)^\top \Sigma_c^{-1} (\mu_c - \mathbf{x}_i)\right]}{ \sum_c \exp\left[-(\mu_c - \mathbf{x}_i)^\top \Sigma_c^{-1} (\mu_c - \mathbf{x}_i)\right] }.
\end{aligned}
\end{equation}

The next step is to update $\mu_c$ to minimize the KL-divergence between $g(\mathbf{x}|c)$ and $p(\mathbf{x}|c)$:
\begin{equation}
    \begin{aligned}
        \mathrm{KL} [ q(\mathbf{x}|c) | g ( \mathbf{x} |c  )   ] & = - \int q(\mathbf{x}|c) \log g( \mathbf{x} |c ) d\mathbf{x} - H[q(\mathbf{x}|c)] \\
        &= -\frac{1}{n_c} \int \sum_{i\in S_c} \mathcal{N}(\mathbf{x};\mathbf{x}_i,\epsilon^2) \log g( \mathbf{x} |c ) d\mathbf{x} - H[q(\mathbf{x}|c)] \\ 
        & \approx -\frac{1}{n_c} \sum_{i\in S_c } \log g( \mathbf{x}_i |c ) - H[q(\mathbf{x}|c)] 
    \end{aligned}
\end{equation}
Minimizing the above w.r.t.\ $\mu_c$ gives:
\begin{equation}
    \mu^{(t)}_c = \frac{1}{N} \sum_{i=1}^N q(c|i) x_i = \sum_{i=1}^N \frac{\log q^{(t-1)}(c)}{N \det \Sigma_c} \frac{\exp\left[-(\mu_c - \mathbf{x}_i)^\top \Sigma_c^{-1} (\mu_c - \mathbf{x}_i)\right]}{ \sum_c \exp\left[-(\mu_c - \mathbf{x}_i)^\top \Sigma_c^{-1} (\mu_c - \mathbf{x}_i)\right]} \mathbf{x}_i.
\end{equation}

By properly re-arranging the above terms and writing them into a compact matrix form,  the relationship between the IB approach and self-attention would become clearer. Assume $\Sigma_c = \Sigma  $ is shared across all the clusters. Assume       $\mu_c$ are   normalized w.r.t. $\Sigma_c^{-1}$, i.e., $\mu_c^\top \Sigma_c^{-1} \mu_c = 1$. 
\begin{equation}
\label{eqn:ib_mu_update}
    \mu^{(t)}_c = \sum_{i=1}^N \frac{\log [n_c / N] }{N \det \Sigma} \frac{\exp\left[ \frac{\mu_c^\top \Sigma^{-1} \mathbf{x}_i}{1/2}   \right]}{ \sum_c \exp\left[ \frac{\mu_c^\top \Sigma^{-1} \mathbf{x}_i}{1/2}  \right]} \mathbf{x}_i.
\end{equation}

Define $Z = [{\mu_1^{(t)}}^\top;  \ldots;  {\mu_N^{(t)}}^\top ], V =[\mathbf{x}_1,\ldots,\mathbf{x}_N] W_V, K = [\mu_1^{(t-1)}, \ldots, \mu_N^{(t-1)}], Q = \Sigma^{-1} [\mathbf{x}_1,\ldots, \mathbf{x}_N]$. Define $d=1/2$. Then the above update \eqref{eqn:ib_mu_update} can be written as:
\begin{equation}
    Z = \mathrm{Softmax}\left( \frac{Q^\top K}{d} \right) V.
\end{equation}
Here the softmax normalization is applied along the row direction. Thus we conclude the proof for Proposition \ref{prop:ib_sa}.

Proposition \ref{prop:ib_sa} can be proved by following the above road map.

\subsection{Implementation details}  
\paragraph{Architecture implementation} When comparing to other state-of-the-art methods, we add in a depthwise convolution layer in the MLP block following the practice in previous methods \cite{xie2021segformer, el2021xcit}. For the shortcut connection, we multiply the residual path with a learnable parameter to stabilize the training, following the same practice in \cite{touvron2021going}.

\label{app:implementation_details}
\paragraph{ImageNet classification} For all the experiments and ablation studies, the models are pretrained on ImageNet-1K if not specified additionally. The training recipes   follow the one used in \cite{touvron2021training} for both the baseline model and our proposed FAN model family.
Specifically, 
we train FAN for 300
epochs using AdamW  with a learning rate of 2e-3.
We use 5 epochs to linearly warmup the model. We adopt a cosine decaying
schedule afterward. We use a batch size of 2048 and a
weight decay of 0.05. We adopt the same data augmentation schemes as \cite{touvron2021training}  including Mixup, Cutmix, RandAugment, and Random Erasing. 
We use Exponential Moving Average (EMA) to speed up the model convergence in a similar manner as timm library \cite{rw2019timm}. For the image classification tasks, we also include two class attention blocks at the top layers as proposed by \citeauthor{touvron2021going}.

\paragraph{Semantic segmentation and object detection} For FAN-ViT, we follow the same decoder proposed in semantic transformer (SETR) \cite{zheng2021rethinking} and the same training setting used in Segformer \cite{xie2021segformer}. For object detection, we finetune the faster RCNN \cite{ren2015faster} with 2x multi-scale training. The resolution of the training image is randomly selected from 640$\times$640 to 896 $\times$ 896. We use a deterministic image resolution of size 896$\times$ 896 for testing.  

For FAN-Swin and FAN-Hybrid, We finetune Mask R-CNN \cite{he2017mask}  on the
COCO dataset. Following Swin
Transformer \cite{liu2021swin}, we use multi-scale training, AdamW optimizer, and 3x schedule. The codes are developed using MMSegmentation \cite{contributors2020mmsegmentation} and MMDetection \cite{chen2019mmdetection} toolbox.

\paragraph{Corruption dataset preparation}

For ImageNet-C, we directly download it from the mirror image provided by \citeauthor{hendrycks2019benchmarking}. For Cityscape-C and COCO-C, we follow \citeauthor{kamann2020benchmarking} and generate 16  algorithmically generated corruptions from noise, blur, weather and digital categories. 

\paragraph{Evaluation metrics} For ImageNet-C, we use \textit{retentaion} as a main metric to measure the robustness of the model which is defined as $\frac{\text{ImageNet-C Acc.}}{\text{ImageNet Clean Acc}}$. It measures how much accuracy can be reserved when evaluated on ImageNet-C dataset. When comparing with other models, we also report the mean corruption error (mCE) in the same manner defined in the ImageNet-C paper \cite{hendrycks2019benchmarking}. The evaluation code is based on timm library \cite{rw2019timm}. For semantic segmentation and object detection, we load the ImageNet-1k pretrained weights and finetune on Cityscpaes and COCO clean image dataset. Then we directly evaluate the performance on Cityscapes-C and COCO-C. We report
semantic segmentation performance using mean Intersection over Union (mIoU) and object detection performance using mean average precision (mAP). 

\subsection{Impact of head numbers}
\begin{figure}[t] 
\small
\centering
    \begin{minipage}[c]{\linewidth}
    \tiny
    \centering
    \begin{overpic}[width=0.6\textwidth]{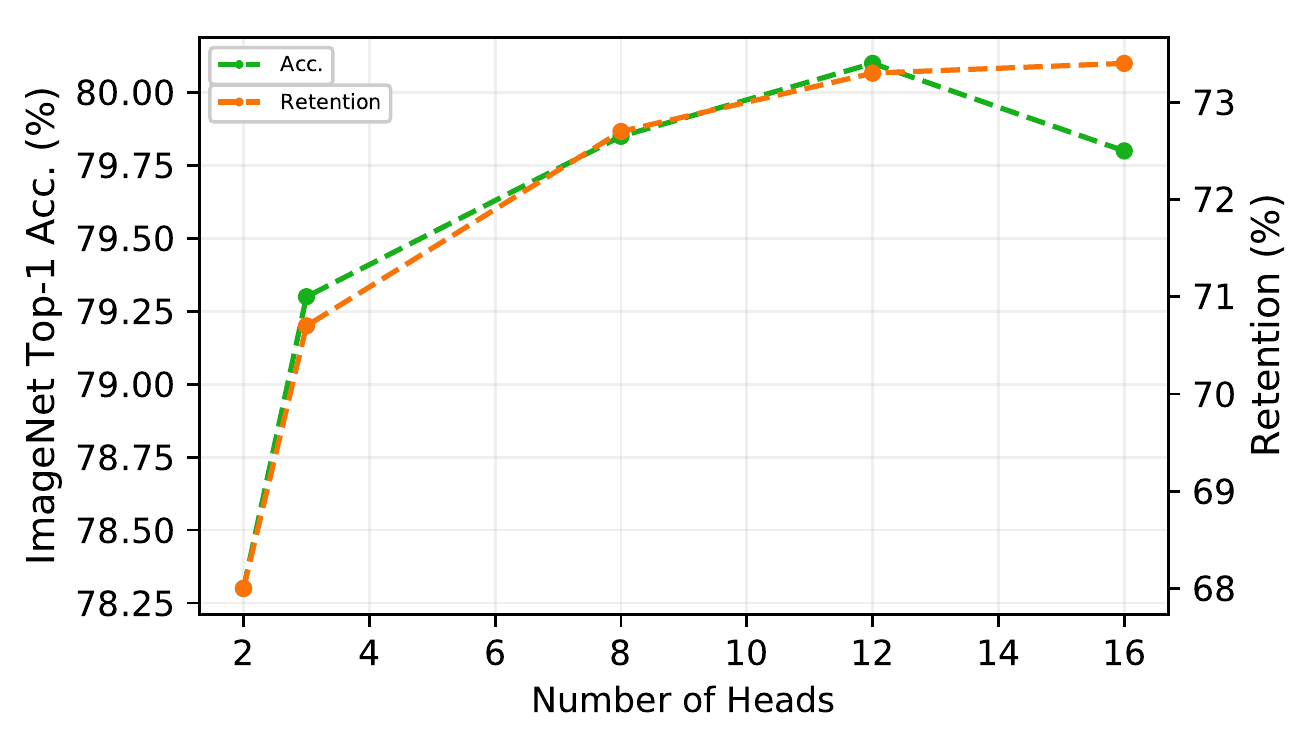}
    \put(35,20){
    \tiny
    \setlength\tabcolsep{0.5mm}
    \renewcommand{\arraystretch}{1}
    \begin{tabular}{l|ccc} 
      Model  & \#Heads. & Clean / Robust  \\ \hline
      DeiT-S (\citeauthor{touvron2021training})  & 2  & 78.3 / 68.0 \\
      DeiT-S (\citeauthor{touvron2021training})  & 3  & 79.3 / 70.7 \\
      DeiT-S (\citeauthor{touvron2021training})  & 8  & 79.9 / 72.7 \\
      DeiT-S (\citeauthor{touvron2021training})  & 12  & 80.1 / 73.3 \\
      DeiT-S (\citeauthor{touvron2021training})  & 16  & 79.8 / 73.4 \\
    \end{tabular}}
    \end{overpic}
    \end{minipage}\hfill
    
\caption{\textbf{Impacts of head number on model robustness.} 
}
\label{fig:head_ablation}
\end{figure}

\subsection{Detailed benchmark results on corrupted images on classification, segmentation and detection}
The by category robustness of selected models and FAN models are shown in Tab. \ref{tab:robustness_imc}, Tab. \ref{tab:benchmark_cityscapes} and Tab. \ref{tab:robustness_coco} respectively. As shown, the strong robustness of FAN is transferrable to all downstreaming tasks.

\input{tables/robustness_imagenet_c}

\input{tables/robustness_cityscape_c}

\input{tables/robustness_coco_c}

\vspace{-2mm}
\subsection{Architecture details of FAN-Swin and FAN-Hybrid}

For FAN-Swin architecture, we follow the same macro architecture design by only replacing the conventional self-attention module with the efficient window shift self-attention in the same manner as proposed in the Swin transformer \cite{liu2021swin}. For the FAN-Hybrid architecture, we use three convolutional building blocks for each stage in the same architecture as proposed in ConvNeXt \cite{liu2022convnet}.

\vspace{-2mm}
\subsection{Feature clustering and visualizations}
\begin{figure*}[t] 
\centering
\includegraphics[width=1.0\linewidth]{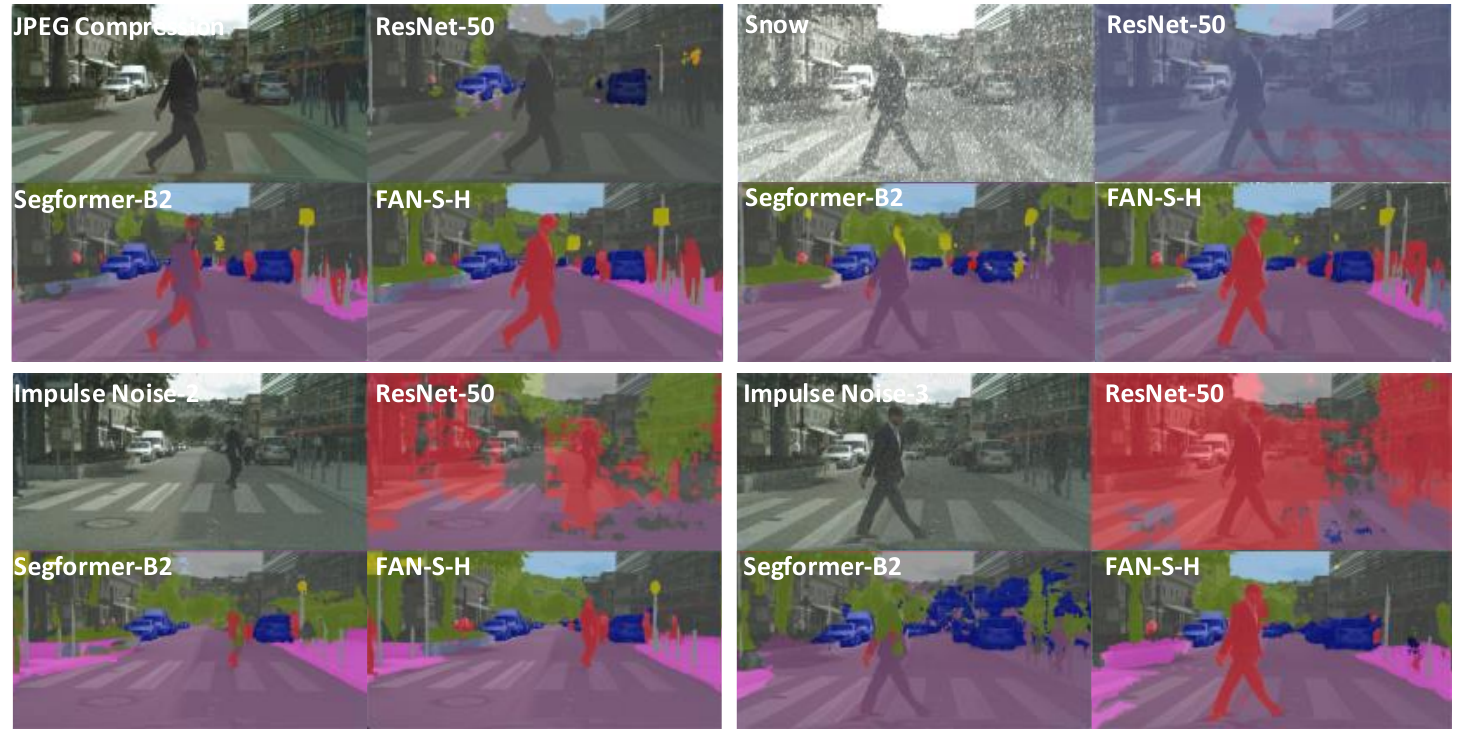}
\caption{\footnotesize Visualization on Cityscapes. A video demonstration is available with \href{https://drive.google.com/file/d/1A9Roe7SanJV145qh481nDpf5JAmiJ-Fl/view?usp=sharing}{external player}
}
\vspace{-3mm}
\label{fig:hook}
\end{figure*}

To cluster the token features, we first normalize the tokens taken from the second last block's output with a SoftMax function. We then calculate a self-correlation matrix based on the normalized tokens and use it as the affinity matrix for spectral clustering. Figure \ref{fig:all} provides more visualization on clustering results of token features from our FAN, ViT and CNN models. The visualization on Cityscape is shown in Figure \ref{fig:hook}.

\input{figures/mask_vis}

%% file: tables/robustness_imagenet_c.tex
\begin{table*}[h]
\begin{savenotes}
\begin{minipage*}{1.0\textwidth}
\centering
% \captionsetup{font={footnotesize}}
\caption{\textbf{Coomparison of model robustness on ImageNet-C (\%)}. FAN shows stronger robustness than other models under all the image corruption settings. `ResNet-50$^*$' denotes our reproduced results with the same training and augmentation recipes for fair comparison.}
\label{tab:robustness_imc}
\vspace{2mm}
% \vspace{-4pt}
% \vspace{-3mm}
\setlength{\tabcolsep}{0.5mm}
% \vspace{-5pt}
\resizebox{\textwidth}{!}{

\addtolength{\tabcolsep}{2pt}
\begin{tabular}{l|c|c|cccc|cccc|cccc|cccc}
% \toprule
\multirow{2}{*}{Model} & \multirow{2}{*}{Param. } & \multirow{2}{*}{Average } & \multicolumn{4}{c|}{Blur} & \multicolumn{4}{c|}{Noise} & \multicolumn{4}{c|}{Digital} & \multicolumn{4}{c}{Weather} \\
\cline{4-19}
& &  & Motion & Defoc & Glass & \multicolumn{1}{c|}{Gauss} & Gauss & Impul & Shot & \multicolumn{1}{c|}{Speck} & Contr & Satur & JPEG & \multicolumn{1}{l|}{Pixel} & \multicolumn{1}{l}{Bright.} & \multicolumn{1}{l}{Snow} & \multicolumn{1}{l}{Fog} & \multicolumn{1}{l}{Frost} \\
\toprule
\multicolumn{18}{c}{Mobile Setting ($<$ 10M)}  \\
% \bottomrule
% \midrule
\toprule
\multicolumn{1}{l|}{ResNet-18 (\citeauthor{he2016deep})} & 11M & \multicolumn{1}{c|}{32.7 } & 29.6 & 28.0 & 22.9 & \multicolumn{1}{c|}{32.0} & 22.7 & 17.6 & 20.8 & \multicolumn{1}{c|}{27.7} & 30.8 & 52.7 & 46.3 & \multicolumn{1}{c|}{42.3} & 58.8 & 24.1 & 41.7 & 28.2 \\
\multicolumn{1}{l|}{MobileNetV2 (\citeauthor{sandler2018mobilenetv2})} & \pzo4M &  \multicolumn{1}{c|}{35.0} & 33.4 & 29.6 & 21.3 & \multicolumn{1}{c|}{32.9} & 24.4 & 21.5 & 23.7 & \multicolumn{1}{c|}{32.9} & 57.6 & 49.6 & 38.0 & \multicolumn{1}{c|}{62.5} & 28.4 & 45.2 & 37.6 & 28.3 \\
\multicolumn{1}{l|}{EfficientNet-B0 (\citeauthor{tan2019efficientnet})} & \pzo5M & \multicolumn{1}{c|}{41.1 } & 36.4 & 26.8 & 26.9 & \multicolumn{1}{c|}{39.3} & 39.8 & 38.1 & 47.1 & \multicolumn{1}{c|}{39.9} & 65.2 & 58.2 & 52.1 & \multicolumn{1}{c|}{69.0} & 37.3 & 55.1 & 44.6 & 37.4 \\
%
% \midrule
\multicolumn{1}{l|}{PVT-V2-B0 (\citeauthor{wang2021pyramid})} & \pzo3M & \multicolumn{1}{c|}{36.2} & 30.8 & 24.9 & 34.0 & \multicolumn{1}{c|}{35.8} & 33.1 & 35.2 & 44.2 & \multicolumn{1}{c|}{50.6} & 59.3 & 50.8 & 36.6 & \multicolumn{1}{c|}{61.9} & 38.6 & 50.7 & 45.9 & 41.8 \\
\multicolumn{1}{l|}{PVT-V2-B1 (\citeauthor{wang2021pyramid})} & 13M &  \multicolumn{1}{c|}{51.7 } & 45.7 & 41.3 & 30.5 & \multicolumn{1}{c|}{43.9} & 48.1 & 46.2 & 46.6 & \multicolumn{1}{c|}{55.0} & 57.6 & 68.6 & 59.9 & \multicolumn{1}{c|}{50.2} & 71.0 & 49.8 & 56.8 & 53.0 \\
\midrule
\multicolumn{1}{l|}{FAN-T-ViT-P16 (Ours)}& \pzo7M & \multicolumn{1}{c|}{\textbf{57.5 }} & {52.4}  & \textbf{48.3}  & \textbf{37.4}  & \multicolumn{1}{c|}{\textbf{51.5}} & {54.8}  & {54.7}  & {53.1}  & \multicolumn{1}{c|}{{60.2}} & \textbf{66.6}  & {72.8}  & {62.7}  & \multicolumn{1}{c|}{{56.7}} & {74.3}  & \textbf{55.5}  & \textbf{61.4}  & \textbf{53.6}  \\
%
% \multicolumn{1}{l|}{FAN-T-ViT-P8 (Ours)}& \pzo8M & \multicolumn{1}{c|}{\textbf{59.3} } & \textbf{53.3}  & \textbf{47.1}  & \textbf{38.7}  & \multicolumn{1}{c|}{\textbf{57.0}} & \textbf{57.0}  & \textbf{57.8}  & \textbf{55.9}  & \multicolumn{1}{c|}{\textbf{62.3}} & \textbf{67.7}  & \textbf{70.2}  & \textbf{64.1}  & \multicolumn{1}{c|}{\textbf{54.5}} & \textbf{74.9}  & \textbf{60.6}  & \textbf{64.6}  & \textbf{59.0}  \\
\multicolumn{1}{l|}{FAN-T-Hybrid-P16 (Ours)}& \pzo8M & \multicolumn{1}{c|}{{57.4 }} & \textbf{52.6}  & {46.7}  & {34.3}  & \multicolumn{1}{c|}{{50.3}} & \textbf{55.5}  & \textbf{55.8}  & \textbf{54.5}  & \multicolumn{1}{c|}{\textbf{61.4}} & {65.8}  & \textbf{73.3}  & \textbf{63.8}  & \multicolumn{1}{c|}{\textbf{47.9}} & \textbf{74.5}  & {55.0}  & {61.4}  & {52.8}  \\
% \bottomrule
\midrule
% %
%
\multicolumn{18}{c}{GPU Setting (20M+)}  \\
\toprule
\multicolumn{1}{l|}{ResNet-50$^*$ (\citeauthor{he2016deep})} & 25M & \multicolumn{1}{c|}{50.6} & 42.1 & 40.1 & 27.2 & \multicolumn{1}{c|}{42.2} & 42.2 & 36.8 & 41.0 & \multicolumn{1}{c|}{50.3} & 51.7 & 69.2 & 59.3 & \multicolumn{1}{c|}{51.2} & 71.6 & 38.5 & 53.9 & 42.3 \\
%
% \midrule
\multicolumn{1}{l|}{ViT-S (\citeauthor{dosovitskiy2020image})} & 22M & \multicolumn{1}{c|}{54.2} & 49.7 & 45.2 & 38.4 & \multicolumn{1}{c|}{48.0} & 50.2 & 47.6 & 49.0 & \multicolumn{1}{c|}{57.5} & 58.4 & 70.1 & 61.6 & \multicolumn{1}{c|}{57.3} & 72.5 & 51.2 & 50.6 & 57.0 \\
% \midrule
\multicolumn{1}{l|}{DeiT-S (\citeauthor{touvron2021training})}& 22M &  \multicolumn{1}{c|}{58.1} 
& 52.6 & 48.9 & {38.1} & \multicolumn{1}{c|}{51.7} & 57.2 & {55.0} & {54.7} & \multicolumn{1}{c|}{{60.8}} & {63.7} & 71.8 & 64.0 & \multicolumn{1}{c|}{58.3} & 73.6 & {55.1} & 61.1 & {60.7} \\
% \midrule
% \multicolumn{1}{l|}{ResNet-18 (\citeauthor{he2016deep})} & \multicolumn{1}{c|}{69} & 29.63 & 28.0 & 22.9 & \multicolumn{1}{c|}{32.0} & 22.7 & 17.6 & 20.8 & \multicolumn{1}{c|}{27.7} & 30.8 & 52.7 & 46.3 & \multicolumn{1}{c|}{42.3} & 58.8 & 24.1 & 33.9 & 28.2 \\
% \midrule
\midrule
\multicolumn{1}{l|}{FAN-S-ViT (Ours)}& 28M & \multicolumn{1}{c|}{{64.5}} & \textbf{61.4}  & \textbf{56.3}  & \textbf{45.6}  & \multicolumn{1}{c|}{\textbf{58.7}} & {62.1}  & {63.0}  & {61.1}  & \multicolumn{1}{c|}{{67.1}} & \textbf{70.9}  & {77.1}  & \textbf{69.4}  & \multicolumn{1}{c|}{\textbf{63.5}} & {78.4}  & \textbf{63.5}  & \textbf{68.2}  & {61.2}  \\
\multicolumn{1}{l|}{FAN-S-Hybrid (Ours)}& 26M & \multicolumn{1}{c|}{\textbf{64.7} } & {60.8}  & {56.0}  & {44.5}  & \multicolumn{1}{c|}{{58.6}} & \textbf{65.6}  & \textbf{66.2}  & \textbf{64.8}  & \multicolumn{1}{c|}{\textbf{69.7}} & {67.5}  & \textbf{77.4}  & {68.7}  & \multicolumn{1}{c|}{{61.0}} & \textbf{78.4}  & {63.2}  & {66.1}  & \textbf{62.4}  \\
% \multicolumn{1}{l|}{FAN-S-ViT-P8 (Ours)}& 29M & \multicolumn{1}{c|}{\textbf{64.5} } & \textbf{61.3}  & \textbf{54.1}  & \textbf{42.8}  & \multicolumn{1}{c|}{\textbf{57.8}} & \textbf{62.5}  & \textbf{63.6}  & \textbf{61.8}  & \multicolumn{1}{c|}{\textbf{68.0}} & \textbf{70.8}  & \textbf{77.1}  & \textbf{68.7}  & \multicolumn{1}{c|}{\textbf{62.5}} & \textbf{78.3}  & \textbf{65.4}  & \textbf{70.1}  & \textbf{62.9}  \\
\midrule

\multicolumn{18}{c}{GPU Setting (50M+)}  \\
\toprule
\multicolumn{1}{l|}{ResNet-101 (\citeauthor{he2019bag})} & 45M & \multicolumn{1}{c|}{59.2 } & 57.0 & 51.9 & 35.6 & \multicolumn{1}{c|}{55.0} & 51.9 & 51.2 & 51.2 & \multicolumn{1}{c|}{61.2} & 67.8 & 75.5 & 67.3 & \multicolumn{1}{c|}{59.9} & 53.6 & 66.2 & 66.4 & 56.4 \\
%
% \midrule
\multicolumn{1}{l|}{ViT-B$^*$ (\citeauthor{dosovitskiy2020image})} & 88M &  \multicolumn{1}{c|}{59.7}    & 60.2 & 55.6 & 50.0 & \multicolumn{1}{c|}{57.6} & 54.9 & 52.9 & 53.2 & 62.0 & \multicolumn{1}{c}{52.3} & 71.5 & 68.7 & 71.7 & \multicolumn{1}{c}{74.9} & 52.8 & 57.1 & 41.7  \\
\multicolumn{1}{l|}{DeiT-B (\citeauthor{touvron2021training})} & 89M & \multicolumn{1}{c|}{62.7} & 56.7 & 52.2 & 43.6 & \multicolumn{1}{c|}{55.1} & 64.9 & 63.5 & 61.2 & \multicolumn{1}{c|}{65.7} & 68.2 & 74.6 & 66.9 & \multicolumn{1}{c|}{61.7} & 76.2 & 59.7 & 68.2 & 64.9 \\
\multicolumn{1}{l|}{Swin-S (\citeauthor{liu2021swin})} & 50M & \multicolumn{1}{c|}{60.4} & 56.7 & 51.4 & 34.8 & \multicolumn{1}{c|}{53.4} & 60.07 & 58.4 & 57.8 & \multicolumn{1}{c|}{62.3} & 65.9 & 73.8 & 66.4 & \multicolumn{1}{c|}{62.4} & 76.0 & 55.9 & 67.4 & 60.7 \\
\midrule
% \multicolumn{1}{l|}{FAN-B-ViT (Ours)}& 54M &  \multicolumn{1}{c|}{\textbf{65.2} } & \textbf{61.2}  & \textbf{55.3}  & \textbf{45.9}  & \multicolumn{1}{c|}{\textbf{57.5}} & \textbf{61.7}  & \textbf{62.8}  & \textbf{60.6}  & \multicolumn{1}{c|}{\textbf{73.1}} & \textbf{71.5}  & \textbf{77.3}  & \textbf{70.0}  & \multicolumn{1}{c|}{\textbf{67.3}} & \textbf{78.35}  & \textbf{63.0}  & \textbf{70.0}  & \textbf{62.9}  \\
%
\multicolumn{1}{l|}{FAN-B-ViT (Ours)}& 54M &  \multicolumn{1}{c|}{{67.0} } & {64.2}  & {58.4}  & {49.7}  & \multicolumn{1}{c|}{{60.8}} & {66.0}  & {67.3}  & {65.0}  & \multicolumn{1}{c|}{{69.8}} & {72.9}  & {78.1}  & {71.2}  & \multicolumn{1}{c|}{{66.9}} & {79.3}  & {64.5}  & {70.9}  & {62.8}  \\
\multicolumn{1}{l|}{FAN-B-Hybrid (Ours)}& 50M &  \multicolumn{1}{c|}{{66.4} } & {62.5}  & {58.0}  & {47.2}  & \multicolumn{1}{c|}{{60.9}} & {67.6}  & {67.9}  & {67.1}  & \multicolumn{1}{c|}{{71.2}} & {70.8}  & {78.0}  & {69.3}  & \multicolumn{1}{c|}{{62.1}} & {78.9}  & {64.8}  & {69.8}  & {63.3}  \\
\multicolumn{1}{l|}{FAN-B-Hybrid-IN22K (Ours)}& 50M &  \multicolumn{1}{c|}{\textbf{70.5} } & \textbf{67.4}  & \textbf{62.9}  & \textbf{55.6}  & \multicolumn{1}{c|}{\textbf{65.4}} & \textbf{70.3}  & \textbf{71.6}  & \textbf{70.1}  & \multicolumn{1}{c|}{\textbf{73.8}} & \textbf{74.1}  & \textbf{79.8}  & \textbf{74.3}  & \multicolumn{1}{c|}{\textbf{79.8}} & \textbf{81.0}  & \textbf{70.2}  & \textbf{72.2}  & \textbf{65.4}  \\
% \midrule
% \multicolumn{18}{c}{FAN finetuned with corruption aware augmentation}  \\
% \bottomrule
%
% \multicolumn{1}{l|}{QualNet18 (\citeauthor{he2016deep})} & \multicolumn{1}{c|}{} & 29.6 & 28.0 & 22.9 & \multicolumn{1}{c|}{32.0} & 22.7 & 17.6 & 20.8 & \multicolumn{1}{c|}{27.7} & 30.8 & 52.7 & 46.3 & \multicolumn{1}{c|}{42.3} & 58.8 & 24.1 & 41.7 & 28.2 \\
% %
% %
% \multicolumn{1}{l|}{QualNet50 (\citeauthor{he2016deep})} & \multicolumn{1}{c|}{} & 29.6 & 28.0 & 22.9 & \multicolumn{1}{c|}{32.0} & 22.7 & 17.6 & 20.8 & \multicolumn{1}{c|}{27.7} & 30.8 & 52.7 & 46.3 & \multicolumn{1}{c|}{42.3} & 58.8 & 24.1 & 41.7 & 28.2 \\
% %
% %
% \multicolumn{1}{l|}{QualNeXt101 (\citeauthor{})} & \multicolumn{1}{c|}{} & 29.6 & 28.0 & 22.9 & \multicolumn{1}{c|}{32.0} & 22.7 & 17.6 & 20.8 & \multicolumn{1}{c|}{27.7} & 30.8 & 52.7 & 46.3 & \multicolumn{1}{c|}{42.3} & 58.8 & 24.1 & 41.7 & 28.2 \\
% \midrule
% %
% %
% \multicolumn{1}{l|}{FAN-S-Finetuned-P16 (Ours)} & \multicolumn{1}{c|}{70.7} & 65.0 & 65.2 & 66.4 & \multicolumn{1}{c|}{66.4} & 67.3 & 68.1 & 67.7 & \multicolumn{1}{c|}{71.6} & 73.1 & 77.2 & 73.1 & \multicolumn{1}{c|}{74.2} & 78.0 & 71.3 & 74.5 & 70.7 \\
% \midrule
% %
% %
% \multicolumn{1}{l|}{FAN-S-Finetuned-P8 (Ours)} & \multicolumn{1}{c|}{72.5} & 67.0 & 65.2 & 67.1 & \multicolumn{1}{c|}{67.0} & 70.3 & 71.3 & 70.6 & \multicolumn{1}{c|}{74.7} & 74.0 & 78.3 & 74.0 & 76.1 & \multicolumn{1}{c|}{79.3} & 73.9 & 76.0 & 73.1  \\
% \midrule
%
\midrule
\multicolumn{18}{c}{GPU Setting (80M+)}  \\
\toprule
\multicolumn{1}{l|}{DeiT-B (\citeauthor{touvron2021training})} & 86M & \multicolumn{1}{c|}{59.7 } & 60.22 & 55.6 & 50.0 & \multicolumn{1}{c|}{57.6} & 54.9 & 52.9 & 53.2 & \multicolumn{1}{c|}{62.0} & 52.3 & 71.5 & 68.7 & \multicolumn{1}{c|}{71.7} & 74.9 & {52.9} & 57.1 & {54.1} \\
\multicolumn{1}{l|}{Swin-B-IN22k (\citeauthor{liu2021swin})} & 88M & \multicolumn{1}{c|}{68.6 } & 66.1 & 62.1 & 48.2 & \multicolumn{1}{c|}{63.2} & 67.3 & 66.2 & 66.4 & \multicolumn{1}{c|}{70.5} & 71.7 & 77.8 & 73.5 & \multicolumn{1}{c|}{74.0} & 80.3 & {66.2} & 74.0 & {66.9} \\
\multicolumn{1}{l|}{ConvNeXt-B (\citeauthor{liu2022convnet})} & 89M &  \multicolumn{1}{c|}{63.6 } & 59.6 & 52.9 & 39.2 & 55.2 & \multicolumn{1}{c}{{65.5}} & 64.8 & 63.7 & 66.7 & \multicolumn{1}{c}{69.9} & 76.2 & 68.9 & {64.6} & \multicolumn{1}{c}{77.8} & 59.2 & {66.7} & 64.3  \\
%
% \midrule
% \multicolumn{1}{l|}{FAN-S-Swin (Ours)} & 31M & \multicolumn{1}{c|}{59.4 } & 55.1 & 49.3 & 35.9 & \multicolumn{1}{c|}{52.3} & 57.8 & 57.4 & 56.0 & \multicolumn{1}{c|}{62.2} & 66.3 & 74.0 & 65.4 & \multicolumn{1}{c|}{58.7} & 75.7 & 57.7 & 65.8 & 56.4 \\
\midrule
\multicolumn{1}{l|}{FAN-L-ViT (Ours)} & 81M & \multicolumn{1}{c|}{{67.7} } &  {64.6} &  {58.8} &  {49.6} & \multicolumn{1}{c|}{ {61.1}} & {66.8} &  {68.5} &  {65.6} & \multicolumn{1}{c|}{ {70.1}} &  {72.5} &  {78.4} &  {71.3} & \multicolumn{1}{c|}{{69.8}} &  {79.7} &  {66.5} &  {71.5} &  {64.8} \\
\multicolumn{1}{l|}{FAN-L-Hybrid (Ours)} & 77M & \multicolumn{1}{c|}{ {68.3} } &  {65.1} &  {59.2} &  {49.2} & \multicolumn{1}{c|}{ {61.9}} & {70.1} &  {71.1} &  {69.4} & \multicolumn{1}{c|}{ {72.7}} &  {72.4} &  {77.6} &  {71.8} & \multicolumn{1}{c|}{{66.6}} &  {79.6} &  {65.6} &  {71.3} &  {65.7} \\
\multicolumn{1}{l|}{FAN-L-Hybrid-IN22K (Ours)} & 77M & \multicolumn{1}{c|}{\textbf{73.6} } & \textbf{71.2} & \textbf{67.5} & \textbf{58.9} & \multicolumn{1}{c|}{\textbf{69.3}} & \textbf{73.9} & \textbf{75.1} & \textbf{73.4} & \multicolumn{1}{c|}{\textbf{76.6}} & \textbf{76.4} & \textbf{81.6} & \textbf{76.8} & \multicolumn{1}{c|}{{74.0}} & \textbf{82.5} & \textbf{73.6} & \textbf{74.3} & \textbf{69.6} \\
% \midrule
%

% \bottomrule
\end{tabular}
}
\end{minipage*}
\end{savenotes}
% \vspace{-3mm}
\end{table*}

%% file: tables/robustness_cityscape_c.tex
\begin{table*}[h!]
\begin{savenotes}
\begin{minipage*}{1.0\textwidth}
%\begin{table}[h!]
\centering
\captionsetup{font={footnotesize}}
\vspace{-2mm}
\caption{\textbf{Comparison of Model Robustness on Cityscapes-C (\%)}. FAN shows stronger robustness than both CNN and transformer models,  for all the image corruption settings. ``DLv3+'' refer to DeepLabv3+ \cite{chen2018encoder}. The mIoUs of compared CNN models are adopted from \cite{kamann2020benchmarking}. The mIoU of ConvNeXt, DeiT, Swin and SegFormer models are our reproduced results. % \textcolor{red}{[  ConvNext Results? ]}
}
\label{tab:benchmark_cityscapes}
% \vspace{2mm}
\vspace{-4pt}
\setlength{\tabcolsep}{0.3mm}
\resizebox{\textwidth}{!}{
\addtolength{\tabcolsep}{2pt}
\begin{tabular}{l|c|cccc|cccc|cccc|cccc}
% \toprule
\multirow{2}{*}{Model} & \multirow{2}{*}{Average} & \multicolumn{4}{c|}{Blur} & \multicolumn{4}{c|}{Noise} & \multicolumn{4}{c|}{Digital} & \multicolumn{4}{c}{Weather} \\
\cline{3-18}
&  & Motion & Defoc & Glass & \multicolumn{1}{c|}{Gauss} & Gauss & Impul & Shot & \multicolumn{1}{c|}{Speck} & Bright & Contr & Satur & \multicolumn{1}{l|}{JPEG} & \multicolumn{1}{l}{Snow} & \multicolumn{1}{l}{Spatt} & \multicolumn{1}{l}{Fog} & \multicolumn{1}{l}{Frost} \\
\toprule
%\multicolumn{1}{l|}{DLv3+ (MBv2)} & \multicolumn{1}{c|}{72.0} & 53.5 & 49.0 & 45.3 & \multicolumn{1}{c|}{49.1} & 6.4 & 7.0 & 6.6 & \multicolumn{1}{c|}{16.6} & 51.7 & 46.7 & 32.4 & \multicolumn{1}{c|}{27.2} & 13.7 & 38.9 & 47.4 & 17.3 \\
%
\multicolumn{1}{l|}{DLv3+ (R50)} & \multicolumn{1}{c|}{36.8} & 58.5 & 56.6 & 47.2 & \multicolumn{1}{c|}{57.7} & 6.5 & 7.2 & 10.0 & \multicolumn{1}{c|}{31.1} & 58.2 & 54.7 & 41.3 & \multicolumn{1}{c|}{27.4} & 12.0 & 42.0 & 55.9 & 22.8 \\
\multicolumn{1}{l|}{DLv3+ (R101)} & \multicolumn{1}{c|}{39.4} & 59.1 & 56.3 & 47.7 & \multicolumn{1}{c|}{57.3} & 13.2 & 13.9 & 16.3 & \multicolumn{1}{c|}{36.9} & 59.2 & 54.5 & 41.5 & \multicolumn{1}{c|}{37.4} & 11.9 & 47.8 & 55.1 & 22.7 \\
%
% \multicolumn{1}{l|}{DLv3+ (X41)} & \multicolumn{1}{c|}{77.8} & 61.6 & 54.9 & 51.0 & \multicolumn{1}{c|}{54.7} & 17.0 & 17.3 & 21.6 & \multicolumn{1}{c|}{43.7} & 63.6 & 56.9 & 51.7 & \multicolumn{1}{c|}{38.5} & 18.2 & 46.6 & 57.6 & 20.6 \\
\multicolumn{1}{l|}{DLv3+ (X65)} & \multicolumn{1}{c|}{42.7} & 63.9 & 59.1 & 52.8 & \multicolumn{1}{c|}{59.2} & 15.0 & 10.6 & 19.8 & \multicolumn{1}{c|}{42.4} & 65.9 & 59.1 & 46.1 & \multicolumn{1}{c|}{31.4} & 19.3 & 50.7 & 63.6 & 23.8 \\
\multicolumn{1}{l|}{DLv3+ (X71)} & \multicolumn{1}{c|}{42.5} & 64.1 & 60.9 & 52.0 & \multicolumn{1}{c|}{60.4} & 14.9 & 10.8 & 19.4 & \multicolumn{1}{c|}{41.2} & 68.0 & 58.7 & 47.1 & \multicolumn{1}{c|}{40.2} & 18.8 & 50.4 & 64.1 & 20.2 \\
\midrule
\multicolumn{1}{l|}{ICNet (\citeauthor{zhao2018icnet})} & \multicolumn{1}{c|}{28.0} & 45.8 & 44.6 & 47.4 & \multicolumn{1}{c|}{44.7} & 8.4 & 8.4 & 10.6 & \multicolumn{1}{c|}{27.9} & 41.0 & 33.1 & 27.5 & \multicolumn{1}{c|}{34.0} & 6.3 & 30.5 & 27.3 & 11.0 \\
\multicolumn{1}{l|}{FCN8s (\citeauthor{long2015fully})} & \multicolumn{1}{c|}{27.4} & 42.7 & 31.1 & 37.0 & \multicolumn{1}{c|}{34.1} & 6.7 & 5.7 & 7.8 & \multicolumn{1}{c|}{24.9} & 53.3 & 39.0 & 36.0 & \multicolumn{1}{c|}{21.2} & 11.3 & 31.6 & 37.6 & 19.7 \\
\multicolumn{1}{l|}{DilatedNet (\citeauthor{yu2015multi})} & \multicolumn{1}{c|}{30.3} & 44.4 & 36.3 & 32.5 & \multicolumn{1}{c|}{38.4} & 15.6 & 14.0 & 18.4 & \multicolumn{1}{c|}{32.7} & 52.7 & 32.6 & 38.1 & \multicolumn{1}{c|}{29.1} & 12.5 & 32.3 & 34.7 & 19.2 \\
\multicolumn{1}{l|}{ResNet-38} & \multicolumn{1}{c|}{32.6} & 54.6 & 45.1 & 43.3 & \multicolumn{1}{c|}{47.2} & 13.7 & 16.0 & 18.2 & \multicolumn{1}{c|}{38.3} & 60.0 & 50.6 & 46.9 & \multicolumn{1}{c|}{14.7} & 13.5 & 45.9 & 52.9 & 22.2 \\
\multicolumn{1}{l|}{PSPNet (\citeauthor{zhao2017pyramid})} & \multicolumn{1}{c|}{34.5} & 59.8 & 53.2 & 44.4 & \multicolumn{1}{c|}{53.9} & 11.0 & 15.4 & 15.4 & \multicolumn{1}{c|}{34.2} & 60.4 & 51.8 & 30.6 & \multicolumn{1}{c|}{21.4} & 8.4 & 42.7 & 34.4 & 16.2 \\
\multicolumn{1}{l|}{ConvNeXt-T (\citeauthor{liu2022convnet})} & \multicolumn{1}{c|}{54.4} & 64.1 & 61.4 & 49.1 & \multicolumn{1}{c|}{62.1} & 34.9 & 31.8 & 38.8 & \multicolumn{1}{c|}{56.7} & 76.7 & 68.1 & 76.0 & 51.1 & \multicolumn{1}{c}{25.0} & 58.7 & 74.2 & 35.1  \\
\midrule
% \multicolumn{1}{c|}{DLv3+$^*$\footnote{Re-implemented DeepLabv3+ to verify the correctness of our Cityscapes-C evaluation. The performance of the re-implemented DeepLabv3+ is at a similar level with the originally reported results.}} & \multicolumn{1}{c|}{79.8} & 58.1 & 51.9 & 42.0  & \multicolumn{1}{c|}{54.5 } & 8.4  & 14.1  & 12.6  & \multicolumn{1}{c|}{39.2} & 76.4  & 56.0  & 68.3  & \multicolumn{1}{c|}{21.5} & 10.3  & 41.3  & 73.5  & 27.6  \\
\multicolumn{1}{l|}{SETR (DeiT-S) (\citeauthor{zheng2021rethinking})}& \multicolumn{1}{c|}{{55.5}} & {61.8}  & {61.0}  & {59.2}  & \multicolumn{1}{c|}{{62.1}} & {36.4}  & {33.8}  & {42.2}  & \multicolumn{1}{c|}{{61.2}} & {73.1} &{63.8}  & {69.1}   & \multicolumn{1}{c|}{{49.7}}   & {41.2}  & {60.8} & 63.8  & {32.0}  \\
\multicolumn{1}{l|}{Swin-T (\citeauthor{liu2021swin})}& \multicolumn{1}{c|}{{47.5}} & {62.1}  & {61.0}  & {48.7}  & \multicolumn{1}{c|}{{62.2}} & {22.1}  & {24.8}  & {25.1}  & \multicolumn{1}{c|}{{42.2}} & {75.8} & {62.1}  & {75.7}   & \multicolumn{1}{c|}{{{33.7}}}   & {19.9}  & 56.9 & {72.1}  & {30.0}  \\
\multicolumn{1}{l|}{SegFormer-B0 (\citeauthor{xie2021segformer})}& \multicolumn{1}{c|}{{48.9}} & {59.3}  & {58.9}  & {51.0}  & \multicolumn{1}{c|}{{59.1}} & {25.1}  & 26.6 & {30.4}  & {50.7}  & \multicolumn{1}{c}{{73.3}} & 66.3 & {71.9}  & {31.2}    & \multicolumn{1}{c}{{22.1}} & {52.9}  & {65.3}  & {31.2}  \\
\multicolumn{1}{l|}{SegFormer-B1 (\citeauthor{xie2021segformer})}& \multicolumn{1}{c|}{{52.6}} & {63.8}  & {63.5}  & {52.0}  & \multicolumn{1}{c|}{{29.8}} & {23.3}  & 35.4 & {56.2}  & {76.3}  & \multicolumn{1}{c}{{70.8}} & 74.7 & {36.1}  & {56.2}    & \multicolumn{1}{c}{{28.3}} & {60.5}  & {70.5}  & {36.3}  \\%
\multicolumn{1}{l|}{SegFormer-B2 (\citeauthor{xie2021segformer})}& \multicolumn{1}{c|}{{55.8}} & {68.1}  & {67.6}  & {58.8}  & \multicolumn{1}{c|}{{68.1}} & {23.8}  & 23.1 & {27.2}  & {47.0}  & \multicolumn{1}{c}{{79.9}} & 76.2 & {78.7}  & {46.2}    & \multicolumn{1}{c}{{34.9}} & {64.8}  & {76.0}  & {42.1}  \\
%
% \multicolumn{1}{l|}{SegFormer-B3 (\citeauthor{xie2021segformer})}& \multicolumn{1}{c|}{{55.8}} & {68.1}  & {67.6}  & {58.8}  & \multicolumn{1}{c|}{{68.1}} & {23.8}  & 23.1 & {27.2}  & {47.0}  & \multicolumn{1}{c}{{79.9}} & 76.2 & {78.7}  & {46.2}    & \multicolumn{1}{c}{{34.9}} & {64.8}  & {76.0}  & {42.1}  \\
% %
% \multicolumn{1}{l|}{SegFormer-B4 (\citeauthor{xie2021segformer})}& \multicolumn{1}{c|}{{55.8}} & {68.1}  & {67.6}  & {58.8}  & \multicolumn{1}{c|}{{68.1}} & {23.8}  & 23.1 & {27.2}  & {47.0}  & \multicolumn{1}{c}{{79.9}} & 76.2 & {78.7}  & {46.2}    & \multicolumn{1}{c}{{34.9}} & {64.8}  & {76.0}  & {42.1}  \\
% %
% \multicolumn{1}{l|}{SegFormer-B5 (\citeauthor{xie2021segformer})}& \multicolumn{1}{c|}{{55.8}} & {68.1}  & {67.6}  & {58.8}  & \multicolumn{1}{c|}{{68.1}} & {23.8}  & 23.1 & {27.2}  & {47.0}  & \multicolumn{1}{c}{{79.9}} & 76.2 & {78.7}  & {46.2}    & \multicolumn{1}{c}{{34.9}} & {64.8}  & {76.0}  & {42.1}  \\
\midrule
% \multicolumn{1}{l|}{FAN-S-ViT (Ours)}& \multicolumn{1}{c|}{{56.9}} &  {68.2}  &  {68.4}  &  {58.8}  & \multicolumn{1}{c|}{ {68.2}} & {18.2}  & {20.1}  & {18.8}  & \multicolumn{1}{c|}{{38.2}}  & {79.1}  &  {76.1}  & \multicolumn{1}{c}{{76.8}} &   \textbf{63.1} &   {45.2}  & 66.3 & {74.9}  &  {45.9}  \\
% %
% \multicolumn{1}{l|}{FAN-S-Swin (Ours)}& \multicolumn{1}{c|}{{56.5}} & {67.0}  & {65.1}  & {56.3}  & \multicolumn{1}{c|}{{66.2}} & {28.1}  & {32.2}  & {33.1}  & \multicolumn{1}{c|}{{55.4}} & {79.2}  &  {70.1}  & 77.8 & {50.6}  & \multicolumn{1}{c}{{38.2}} &  {69.3}  & {74.3}  & {40.0}    \\
%
\multicolumn{1}{l|}{FAN-T-Hybrid (Ours)}& \multicolumn{1}{c|}{ {57.9}} & {67.1}  & {66.0}  & {57.2}  & \multicolumn{1}{c|}{{66.6}} &  {33.2}  &  {34.3}  &  {36.2}  & \multicolumn{1}{c|}{ {55.6}} &  {80.8} & {72.1}  &  {79.1}  & {54.3}  & \multicolumn{1}{c}{{30.6}}   & {66.1}  &  {78.2}  & {43.8}  \\
\multicolumn{1}{l|}{FAN-S-Hybrid (Ours)}& \multicolumn{1}{c|}{ {66.4}} & {68.6}  & {68.9}  & {61.0}  & \multicolumn{1}{c|}{{70.0}} &  {57.5}  &  {61.3}  &  {62.2}  & \multicolumn{1}{c|}{ {71.5}} &  {80.5} & {74.9}  &  {79.4}  & {62.1}  & \multicolumn{1}{c}{{47.4}}   & {70.8}  &  {77.9}  & {48.8}  \\
\multicolumn{1}{l|}{FAN-B-Hybrid (Ours)}& \multicolumn{1}{c|}{ {67.3}} & {70.0}  & {69.0}  & {64.3}  & \multicolumn{1}{c|}{{70.3}} &  {55.9}  &  {60.4}  &  {61.1}  & \multicolumn{1}{c|}{ {70.9}} &  {81.2} & {76.1}  &  {80.0}  & {57.0}  & \multicolumn{1}{c}{\textbf{54.8}}   & {72.5}  &  {78.4}  & {52.3}  \\
\multicolumn{1}{l|}{FAN-L-Hybrid (Ours)}& \multicolumn{1}{c|}{\textbf{68.5}} & \textbf{70.0}  & \textbf{69.9}  & \textbf{65.3}  & \multicolumn{1}{c|}{\textbf{71.6}} & \textbf{60.0}  & \textbf{64.5}  & \textbf{63.3}  & \multicolumn{1}{c|}{\textbf{71.6}} & \textbf{81.4} & \textbf{76.2}  & \textbf{80.1}  & {62.3}  & \multicolumn{1}{c}{{53.1}}   & \textbf{73.9}  & \textbf{78.9}  & \textbf{54.4}  \\
% \bottomrule
\end{tabular}
}

\end{minipage*}
\end{savenotes}
\vspace{-2mm}
\end{table*}

%% file: tables/robustness_coco_c.tex
\begin{table*}[h!]
\begin{savenotes}
\begin{minipage*}{1.0\textwidth}
\vspace{-1mm}
\caption{\textbf{Comparison of model robustness on COCO-C (\%)}. FAN shows stronger robustness than other models.}
\centering
% \captionsetup{font={footnotesize}}
\label{tab:robustness_coco}
% \vspace{-4pt}
\vspace{2mm}
\setlength{\tabcolsep}{0.5mm}
% \vspace{-5pt}
\resizebox{\textwidth}{!}{

\addtolength{\tabcolsep}{2pt}
\vspace{-2mm}
\begin{tabular}{l|c|cccc|cccc|cccc|cccc}
% \toprule
\multirow{2}{*}{Model} & \multirow{2}{*}{Average} & \multicolumn{4}{c|}{Blur} & \multicolumn{4}{c|}{Noise} & \multicolumn{4}{c|}{Digital} & \multicolumn{4}{c}{Weather} \\
\cline{3-18}
&  & Motion & Defoc & Glass & \multicolumn{1}{c|}{Gauss} & Gauss & Impul & Shot & \multicolumn{1}{c|}{Speck} & Bright. & Contr & Satur & \multicolumn{1}{l|}{JPEG}  & \multicolumn{1}{l}{Snow}& \multicolumn{1}{l}{Spatter}  & \multicolumn{1}{l}{Fog} & \multicolumn{1}{l}{Frost} \\
\toprule
%
% \multicolumn{1}{l|}{ResNet18 (\citeauthor{he2016deep})} & \multicolumn{1}{c|}{32.7} & 29.6 & 28.0 & 22.9 & \multicolumn{1}{c|}{32.0} & 22.7 & 17.6 & 20.8 & \multicolumn{1}{c|}{27.7} & 30.8 & 52.7 & 46.3 & \multicolumn{1}{c|}{42.3} & 58.8 & 24.1 & 41.7 & 28.2 \\
% %
% \multicolumn{1}{l|}{ResNet50$^*$ (\citeauthor{he2016deep})} & \multicolumn{1}{c|}{79} & 42.1 & 40.1 & 27.2 & \multicolumn{1}{c|}{42.2} & 42.2 & 36.8 & 41.0 & \multicolumn{1}{c|}{50.3} & 51.7 & 69.2 & 59.3 & \multicolumn{1}{c|}{51.2} & 71.6 & 38.5 & 53.9 & 42.3 \\
% %
% \midrule
% \multicolumn{1}{l|}{ViT-S (\citeauthor{dosovitskiy2020image})} & \multicolumn{1}{c|}{54.2} & 49.7 & 45.2 & 38.4 & \multicolumn{1}{c|}{48.0} & 50.2 & 47.6 & 49.0 & \multicolumn{1}{c|}{57.5} & 58.4 & 70.1 & 61.6 & \multicolumn{1}{c|}{57.3} & 72.5 & 51.2 & 50.6 & 57.0 \\
% % \midrule
% \multicolumn{1}{l|}{DeiT-S (\citeauthor{touvron2021training})}& \multicolumn{1}{c|}{58.1} 
% & 52.6 & 48.9 & {38.1} & \multicolumn{1}{c|}{51.7} & 57.2 & {55.0} & {54.7} & \multicolumn{1}{c|}{{60.8}} & {63.7} & 71.8 & 64.0 & \multicolumn{1}{c|}{58.3} & 73.6 & {55.1} & 61.1 & {60.7} \\
% \midrule
% \multicolumn{1}{l|}{ResNet-18 (\citeauthor{he2016deep})} & \multicolumn{1}{c|}{69} & 29.63 & 28.0 & 22.9 & \multicolumn{1}{c|}{32.0} & 22.7 & 17.6 & 20.8 & \multicolumn{1}{c|}{27.7} & 30.8 & 52.7 & 46.3 & \multicolumn{1}{c|}{42.3} & 58.8 & 24.1 & 33.9 & 28.2 \\
\multicolumn{1}{l|}{ResNet-50(Faster-RCNN) (\citeauthor{ren2015faster})} & \multicolumn{1}{c|}{21.3} & 16.6 & 18.2 & 11.4 & \multicolumn{1}{c|}{19.9} & 17.1 & 17.3 & 14.0 & \multicolumn{1}{c|}{22.5} & 35.0 & 21.8 & 33.7 & \multicolumn{1}{c|}{  18.2 } & 18.5 & {26.5} & 31.6 & {23.1} \\
% \midrule
\multicolumn{1}{l|}{ResNet-101 (Faster-RCNN) (\citeauthor{ren2015faster})} & \multicolumn{1}{c|}{23.3} & 18.8 & 20.2 &  13.8   & \multicolumn{1}{c|}{22.0} & 19.2 & 16.2 & 19.2 & \multicolumn{1}{c|}{24.4} & 37.1 & 23.7 & 35.7 & 20.0 & \multicolumn{1}{c}{  20.4 } & 28.8 & {33.9} & 24.9  \\
% \midrule
% \multicolumn{1}{l|}{ResNeXt-101(Faster-RCNN (\citeauthor{ren2015faster})} & \multicolumn{1}{c|}{} & 23.6 & 23.7 &     & \multicolumn{1}{c|}{25.2} & 22.0 & 21.0 & 22.3 & \multicolumn{1}{c|}{27.7} & 25.2 & 35.6 & 29.7 & \multicolumn{1}{c|}{   } & 37.0 & {25.9} & 34.4 & {28.4} \\
% \midrule
\midrule
\multicolumn{1}{l|}{SWIN-T (\citeauthor{liu2021swin})} & \multicolumn{1}{c|}{29.3} & 24.2 & 25.8 &18.2     & \multicolumn{1}{c|}{27.8} & 24.6 & 23.7 & 24.8 & \multicolumn{1}{c|}{30.5} & 42.1 & 31.2 & 41.0 & \multicolumn{1}{c|}{ 26.6  } & 26.1 & 36.3 & {40.6} & 30.8 \\
% \midrule

\multicolumn{1}{l|}{DeiT-S (\citeauthor{touvron2021training})} & \multicolumn{1}{c|}{26.9} & 23.6 & 23.7 &  21.8   & \multicolumn{1}{c|}{25.2} & 22.0 & 21.0 & 22.3 & \multicolumn{1}{c|}{27.7} & 37.0 & 25.2 & 35.6 &  \multicolumn{1}{c|}{ 29.7  }  & {25.9} & 31.9 & 34.4 & {28.4} \\
\midrule
% \multicolumn{1}{l|}{FAN-S-ViT (Ours)}& \multicolumn{1}{c|}{{29.0}} & {25.1}  & {22.8}  & {21.4}  & \multicolumn{1}{c|}{{24.5}} & {24.6}  & {25.0}  & {24.8}  & \multicolumn{1}{c|}{{30.2}} & 39.9 &  {28.6}  & {38.7}  &  \multicolumn{1}{c|}{{30.7}}   & {29.3} & 35.0 & {37.5}  & {30.1}  \\
% %
% \multicolumn{1}{l|}{FAN-S-Swin (Ours)} & \multicolumn{1}{c|}{30.7} & 25.6 & 24.4 &  20.6   & \multicolumn{1}{c|}{25.9} & 25.5 & 25.9 & 26.3 & \multicolumn{1}{c|}{32.4} & 43.1 & 32.3 & 41.9 &  \multicolumn{1}{c|}{  28.3 } & \textbf{31.6} & {38.0} & 41.8 & {33.9} \\
%
\multicolumn{1}{l|}{FAN-T-Hybrid (Ours)} & \multicolumn{1}{c|}{{29.7}} & {24.2} & {25.8} & {18.8}     & \multicolumn{1}{c|}{{27.4}} & {23.5} & {22.7} & {24.0} & \multicolumn{1}{c|}{{30.3}}  & 42.4 & {33.5} & {41.2} & \multicolumn{1}{c|}{ {27.9}  } & {28.3} & {36.5} & {41.0} & {32.8} \\
\multicolumn{1}{l|}{FAN-S-Hybrid (Ours)} & \multicolumn{1}{c|}{{35.5}} & {29.1} & {29.5} & {23.5}     & \multicolumn{1}{c|}{{31.5}} & {31.6} & {32.1} & {32.3} & \multicolumn{1}{c|}{{37.1}} & {33.3} & {46.1} & {40.4} & \multicolumn{1}{c|}{ {45.3}  } & {33.5} & {42.4} & {45.9} & {35.5} \\
\multicolumn{1}{l|}{FAN-B-Hybrid (Ours)} & \multicolumn{1}{c|}{{39.0}} & {31.8} & {32.1} & {27.4}     & \multicolumn{1}{c|}{{34.2}} & {34.6} & {35.5} & {35.4} & \multicolumn{1}{c|}{{40.6}} & {50.4} & {43.3} & {49.6} & \multicolumn{1}{c|}{ {36.9}  } & {39.4} & {46.4} & {49.6} & {41.8} \\
\multicolumn{1}{l|}{FAN-L-Hybrid (Ours)} & \multicolumn{1}{c|}{{40.6}} & {33.0} & {33.2} & {28.7}     & \multicolumn{1}{c|}{{35.6}} & {37.1} & {38.0} & {37.5} & \multicolumn{1}{c|}{{42.2}} & {51.3} & {46.0} & {50.3} & \multicolumn{1}{c|}{ {39.2}  } & {41.1} & {47.5} & {50.3} & {43.0} \\
\midrule
\multicolumn{1}{l|}{FAN-B-Hybrid-IN22k (Ours)} & \multicolumn{1}{c|}{{40.6}} & {33.1} & {33.1} & {28.5}     & \multicolumn{1}{c|}{{35.5}} & {36.7} & {38.0} & {37.3} & \multicolumn{1}{c|}{{42.3}} & {51.5} & {45.6} & {50.7} & \multicolumn{1}{c|}{ {39.0}  } & {41.2} & {48.1} & {50.7} & {43.0} \\
\multicolumn{1}{l|}{FAN-L-Hybrid-IN22k (Ours)} & \multicolumn{1}{c|}{\textbf{42.0}}  & \textbf{34.1} & \textbf{34.5}     & \multicolumn{1}{c|}{\textbf{30.8}} & \textbf{37.1} & \textbf{38.5} & \textbf{39.7} & \multicolumn{1}{c|}{\textbf{39.1}} & \textbf{43.6} & \textbf{52.1} & \textbf{47.6} &\textbf{51.3} & \multicolumn{1}{c|}{ \textbf{40.8}  } & \textbf{42.3} & \textbf{49.2} & \textbf{51.8} & \textbf{44.3} \\
%
% \multicolumn{1}{l|}{FAN-S-P8 (Ours)}& \multicolumn{1}{c|}{\textbf{64.5}} & \textbf{61.27}  & \textbf{54.09}  & \textbf{42.8}  & \multicolumn{1}{c|}{\textbf{57.8}} & \textbf{62.5}  & \textbf{63.6}  & \textbf{61.8}  & \multicolumn{1}{c|}{\textbf{68.0}} & \textbf{70.8}  & \textbf{77.1}  & \textbf{68.7}  & \multicolumn{1}{c|}{\textbf{62.5}} & \textbf{78.3}  & \textbf{65.4}  & \textbf{70.1}  & \textbf{62.9}  \\
% \bottomrule 
\end{tabular}
}
\end{minipage*}
\end{savenotes}
\vspace{-1mm}
\end{table*}

%% file: figures/mask_vis.tex
\begin{figure*}
\centering
\setlength{\tabcolsep}{0.5pt}
\begin{tabular}{c ccc ccc c }

Image & FAN &ViT &CNN &Image &FAN &ViT &CNN \\
\toprule
\includegraphics[width=0.123\linewidth]{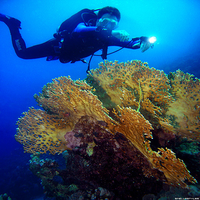} &
\includegraphics[width=0.123\linewidth]{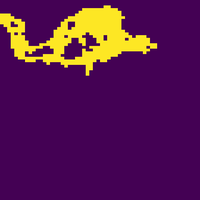} &
\includegraphics[width=0.123\linewidth]{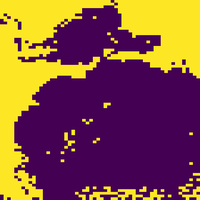} &
\includegraphics[width=0.123\linewidth]{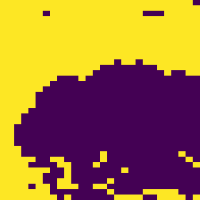} & \quad
\includegraphics[width=0.123\linewidth]{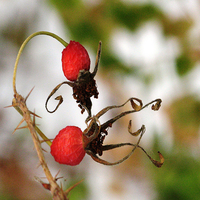} &
\includegraphics[width=0.123\linewidth]{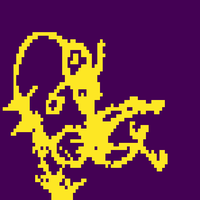} &
\includegraphics[width=0.123\linewidth]{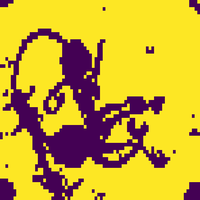} &
\includegraphics[width=0.123\linewidth]{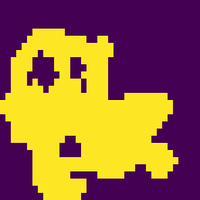}  \\
\includegraphics[width=0.123\linewidth]{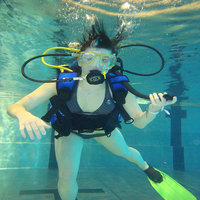} &
\includegraphics[width=0.123\linewidth]{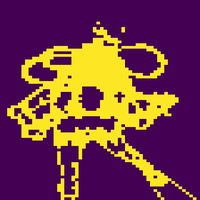} &
\includegraphics[width=0.123\linewidth]{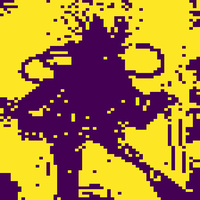} &
\includegraphics[width=0.123\linewidth]{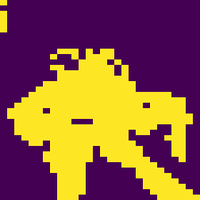} & \quad
\includegraphics[width=0.123\linewidth]{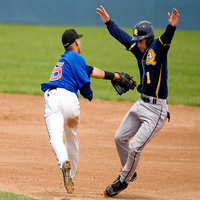} &
\includegraphics[width=0.123\linewidth]{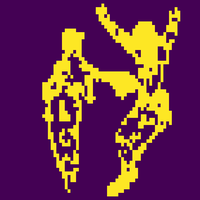} &
\includegraphics[width=0.123\linewidth]{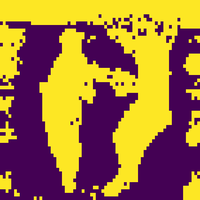} &
\includegraphics[width=0.123\linewidth]{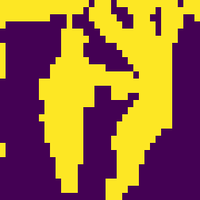}  \\
\includegraphics[width=0.123\linewidth]{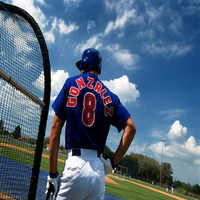} &
\includegraphics[width=0.123\linewidth]{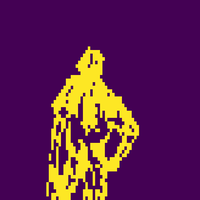} &
\includegraphics[width=0.123\linewidth]{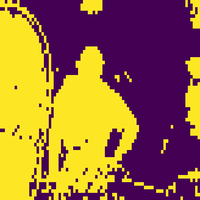} &
\includegraphics[width=0.123\linewidth]{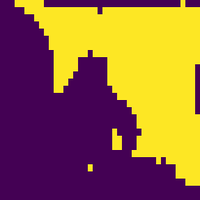} & \quad
\includegraphics[width=0.123\linewidth]{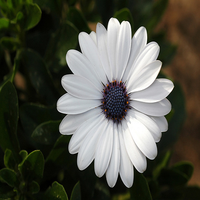} &
\includegraphics[width=0.123\linewidth]{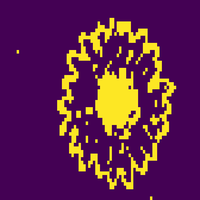} &
\includegraphics[width=0.123\linewidth]{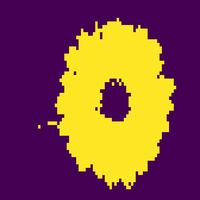} &
\includegraphics[width=0.123\linewidth]{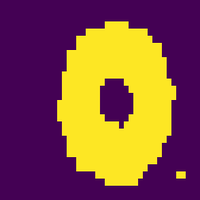}  \\
\includegraphics[width=0.123\linewidth]{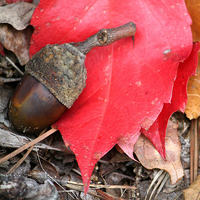} &
\includegraphics[width=0.123\linewidth]{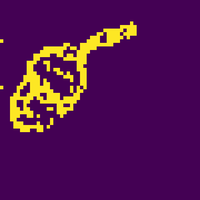} &
\includegraphics[width=0.123\linewidth]{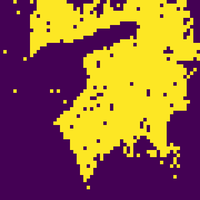} &
\includegraphics[width=0.123\linewidth]{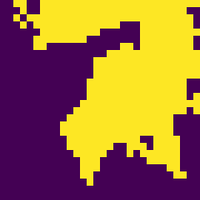} & \quad
\includegraphics[width=0.123\linewidth]{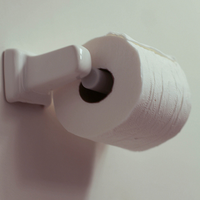} &
\includegraphics[width=0.123\linewidth]{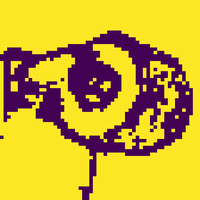} &
\includegraphics[width=0.123\linewidth]{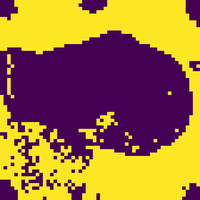} &
\includegraphics[width=0.123\linewidth]{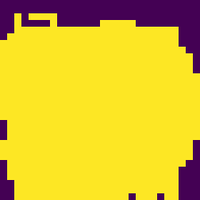}  \\
\includegraphics[width=0.123\linewidth]{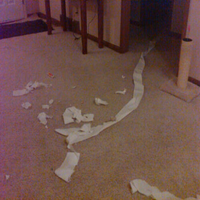} &
\includegraphics[width=0.123\linewidth]{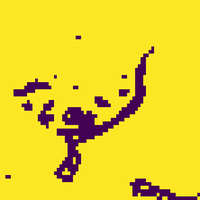} &
\includegraphics[width=0.123\linewidth]{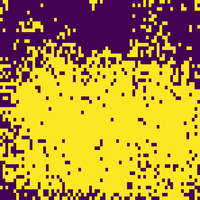} &
\includegraphics[width=0.123\linewidth]{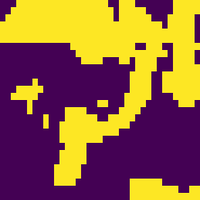} & \quad
\includegraphics[width=0.123\linewidth]{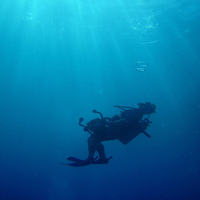} &
\includegraphics[width=0.123\linewidth]{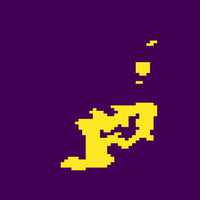} &
\includegraphics[width=0.123\linewidth]{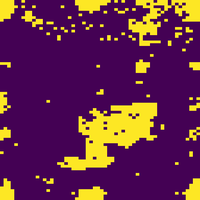} &
\includegraphics[width=0.123\linewidth]{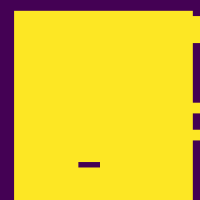}  \\
\includegraphics[width=0.123\linewidth]{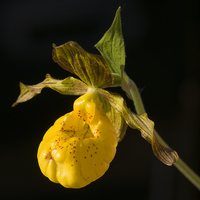} &
\includegraphics[width=0.123\linewidth]{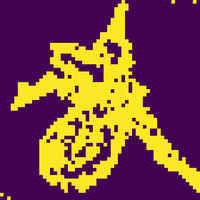} &
\includegraphics[width=0.123\linewidth]{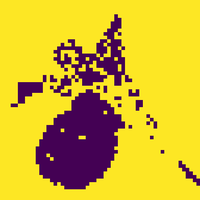} &
\includegraphics[width=0.123\linewidth]{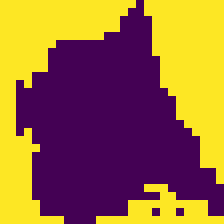} & \quad
\includegraphics[width=0.123\linewidth]{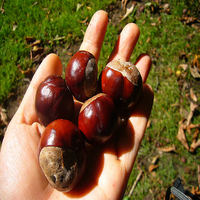} &
\includegraphics[width=0.123\linewidth]{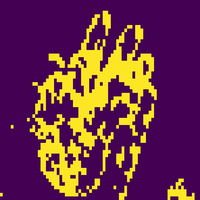} &
\includegraphics[width=0.123\linewidth]{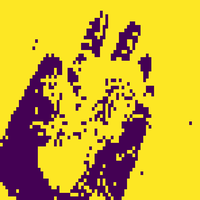} &
\includegraphics[width=0.123\linewidth]{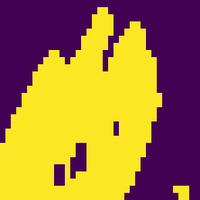}  \\
\includegraphics[width=0.123\linewidth]{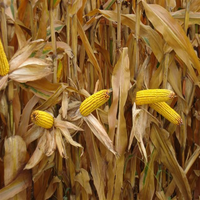} &
\includegraphics[width=0.123\linewidth]{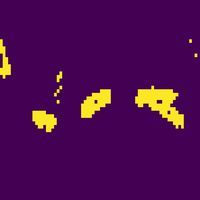} &
\includegraphics[width=0.123\linewidth]{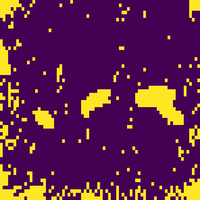} &
\includegraphics[width=0.123\linewidth]{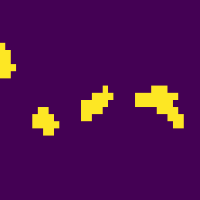} & \quad
\includegraphics[width=0.123\linewidth]{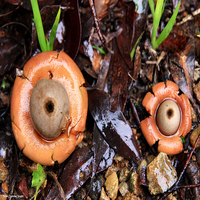} &
\includegraphics[width=0.123\linewidth]{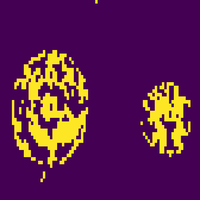} &
\includegraphics[width=0.123\linewidth]{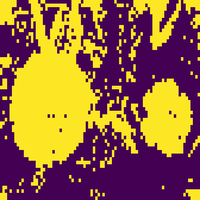} &
\includegraphics[width=0.123\linewidth]{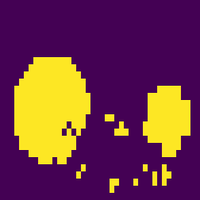} \\
\includegraphics[width=0.123\linewidth]{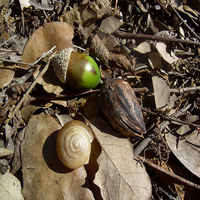} &
\includegraphics[width=0.123\linewidth]{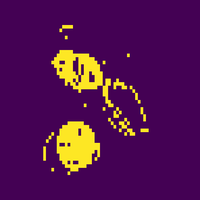} &
\includegraphics[width=0.123\linewidth]{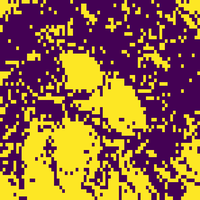} &
\includegraphics[width=0.123\linewidth]{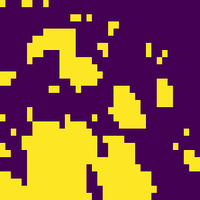} & \quad
\includegraphics[width=0.123\linewidth]{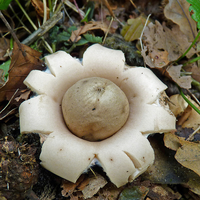} &
\includegraphics[width=0.123\linewidth]{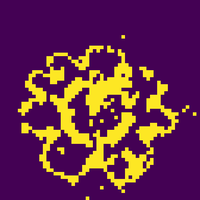} &
\includegraphics[width=0.123\linewidth]{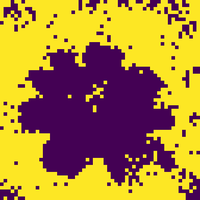} &
\includegraphics[width=0.123\linewidth]{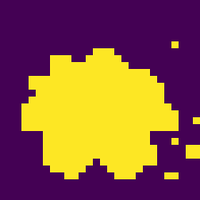}  \\
\includegraphics[width=0.123\linewidth]{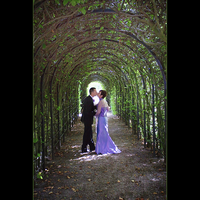} &
\includegraphics[width=0.123\linewidth]{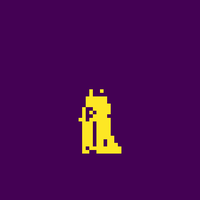} &
\includegraphics[width=0.123\linewidth]{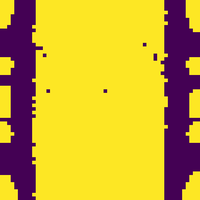} &
\includegraphics[width=0.123\linewidth]{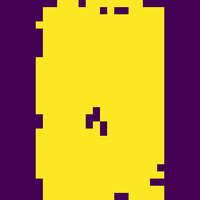} & \quad
\includegraphics[width=0.123\linewidth]{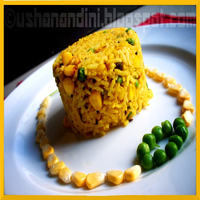} &
\includegraphics[width=0.123\linewidth]{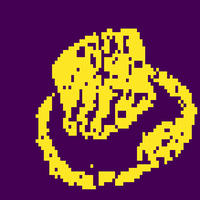} &
\includegraphics[width=0.123\linewidth]{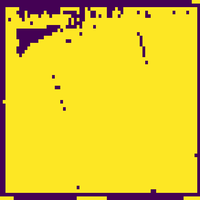} &
\includegraphics[width=0.123\linewidth]{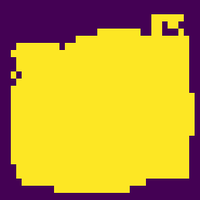} %& \\
\end{tabular}
	\caption{\textbf{clustering visualization.} Our FAN model provides much clearer clusters that feature important regions of foreground objects. }
\label{fig:all}
\end{figure*}